\documentclass[10pt,twocolumn,letterpaper]{article}

\usepackage[pagenumbers]{cvpr} % To force page numbers, e.g. for an arXiv version

\usepackage{multirow}
\usepackage{pifont}
\newcommand{\cmark}{\ding{51}}%
\newcommand{\xmark}{\ding{55}}%

\usepackage{algorithm}
\usepackage{algorithmic}
\usepackage{amsmath}
\usepackage{amssymb}
\usepackage{mathtools}

\usepackage[accsupp]{axessibility}  % Improves PDF readability for those with disabilities.

\usepackage[pagebackref,breaklinks,colorlinks,allcolors=cvprblue]{hyperref}
\hypersetup{
 colorlinks=true,
 linkcolor=red,
 citecolor=cyan,
 filecolor=magenta, 
 urlcolor=cyan,
 }
\def\eg{\emph{e.g.,~}}
\def\ie{\emph{i.e.,~}}
\def\ournet{DriveGEN}

\definecolor{cvprblue}{rgb}{0.21,0.49,0.74}
\usepackage[pagebackref,breaklinks,colorlinks,allcolors=cvprblue]{hyperref}

\def\paperID{445} % *** Enter the Paper ID here
\def\confName{CVPR}
\def\confYear{2025}

\title{DriveGEN: Generalized and Robust 3D Detection in Driving \\via Controllable Text-to-Image Diffusion Generation}

\author{Hongbin Lin$^{1,2}$ \quad Zilu Guo$^{1,2}$ \quad Yifan Zhang$^{3}$ \quad Shuaicheng Niu$^{4}$ \quad \\
Yafeng Li$^{5}$ \quad Ruimao Zhang$^{6}$ \quad Shuguang Cui$^{2,1}$ \quad Zhen Li$^{2,1}$\thanks{Corresponding author.} \\
$^1$FNii-Shenzhen $^2$ SSE, CUHK-Shenzhen $^3$ National University of Singapore \\ $^4$ Nanyang Technological University 
$^5$Baoji University of Arts and Sciences $^6$Sun Yat-sen University \\
} 

\begin{document}
\maketitle

\begin{abstract}
In autonomous driving, vision-centric 3D detection aims to identify 3D objects from images. However, high data collection costs and diverse real-world scenarios limit the scale of training data. Once distribution shifts occur between training and test data, existing methods often suffer from performance degradation, known as Out-of-Distribution (OOD) problems.
% However, high collection costs and diverse real-world scenarios limit the scale of training data. 
% Once the distribution shifts exist between training and test data, existing methods often suffer performance degradation, known as out-of-distribution (OOD) problems.
% existing methods often suffer performance degradation due to distribution shifts between training and test data, known as out-of-distribution (OOD) problems.
% To address the OOD issue, current robust 3D detection methods require extra test-time computation, increasing the energy consumption of vehicles.
% In this paper, we explore controllable Text-to-Image (T2I) diffusion for training data enhancement, thereby improving 3D Detection model robustness.
To address this, controllable Text-to-Image (T2I) diffusion offers a potential solution for training data enhancement, which is required to generate diverse OOD scenarios with precise 3D object geometry.
% Though the controllable Text-to-Image (T2I) diffusion for training data enhancement has been applied for improving 3D Detection model robustness, existing training-based controllable T2I approaches still require substantial labeled data for spatial control.
%
% Considering training-free methods often employ coarse self-attention (SA) features to guide diffusion processes, it struggles to preserve all annotated 3D objects.
% %
%  {\color{red}{(here maybe we should list limitation of prior works instead of our methods)}}
% %
% Nevertheless, existing training-based controllable T2I approaches require substantial labeled data for spatial control, while training-free methods often employ coarse self-attention (SA) features to guide diffusion processes, struggling to preserve all annotated 3D objects.
Nevertheless, existing controllable T2I approaches are restricted by the limited scale of training data or struggle to preserve all annotated 3D objects.
In this paper, we present \textbf{DriveGEN}, a method designed to improve the robustness of 3D detectors in {Driv}ing via Training-Free Controllable Text-to-Image Diffusion {Gen}eration.
% improve detectors across OOD scenarios.
%
% {\color{red}{A sentence to summarize the core contributions of DriverGen to overcome previous drawbacks, later more  }}
%concrete introduction
Without extra diffusion model training, DriveGEN consistently preserves objects with precise 3D geometry across diverse OOD generations, consisting of 2 stages: 
%
% for generalized and robust 3D detection in \textbf{Driv}ing via
% Controllable Text-to-Image Diffusion \textbf{Gen}eration, designed to improve 3D detectors across OOD scenarios even never seen.
% a training-free T2I controllable diffusion method designed to improve the generalizability of 3D detectors across OOD scenarios even never known.
% %
%  {\color{red}{(the model short name should have correspondence)}}
% %
% In practice, {DriveGEN} consists of 2 stages: 
1) Self-Prototype Extraction: We empirically find that self-attention features are semantic-aware but require accurate region selection for 3D objects.
Thus, we extract precise object features via layouts to capture 3D object geometry, termed self-prototypes.
 % {\color{red}{(2 stages to overcome previous drawbacks)}}
% we find the combination of self-attention features and layouts effectively represents object properties like location, shape and orientation during diffusion. Thus, we employ layouts to extract object features of self-attention layers as self-prototypes.
2) Prototype-Guided Diffusion: To preserve objects across various OOD scenarios, we perform semantic-aware feature alignment and shallow feature alignment during denoising.
% with self-prototypes, we conduct feature alignment for object preservation during denoising phase. To retain small objects, we further align noisy latent via time-dependent diffusion features of input images.
Extensive experiments demonstrate the effectiveness of \ournet~in improving 3D detection. The code is available at \href{https://github.com/Hongbin98/DriveGEN}{\textcolor{red}{\emph{Hongbin98/DriveGEN}}}.

% \eg achieving an average of {\textbf{7.6 mAP gain}} across 13 OOD scenarios with only one augmentation.

% introduces more challenging scenarios for 3D vision detectors and significantly improves their generalization capabilities. The code will be released.

% To preserve small objects, we further constrain the diffusion process via low-level feature alignment.

% enhancing detectors with OOD knowledge and significantly improves their generalization capabilities.

% Specifically, \textbf{Free-DriveGen} consists of:
% 1) XXXX
% 2) XXXX

% achieve SOTA even if we don't access any test data，同样结合图一

% 图一里面除了fog再给出更多的未见过的场景
% robust image-based 3d detection
% nus不全部增广，起码试试100 500 1000 2000...

% SA feature map可以是coarse的layout，然后经过object region reweight本质上是获取更准确的layout，也就是self-prototype
% shallow feature WeiJun-paper
% 前面shallow保shape、轮廓低层信息，然后SA alignment是semantic-aware的信息保留
% stage1 的名称可以再修改，layout-based semantic prototype extraction
% 再划清楚一点 每个guidance的特征来源
% 题目要与drive相关才行

\end{abstract}

\vspace{-0.15in}

\section{Introduction}
\label{sec:intro}

Three-dimensional (3D) object detection intends to identify objects and assess their spatial and dimensional attributes via various sensor inputs, which is widely explored in Autonomous Driving (AD) and robotics~\cite{chen20153d,chen2017multi,wang2019pseudo,li2019stereo,yin2021center,chen2023voxelnext,wu2023virtual}.
To save sensor costs, there is a growing trend towards implementing perception systems by vision-centric 3D detection which relies solely on single or multi-view RGB images along with camera calibration information. 
Despite challenges in this field, existing methods~\cite{yang2023bevformer,xu2023mononerd,zhang2023monodetr} have demonstrated promising results across various benchmarks. 
% ~\cite{geiger2012we,caesar2020nuscenes,sun2020scalability}

Behind the success, one prerequisite is that test images share the same distribution as training images. However, this assumption is often not held in diverse real-world scenarios~\cite{lin2022prototype,zhang2023deep,qiu2021source,zhang2020collaborative}, especially for perception systems that are required for long durations, like autonomous vehicles navigating through different weather conditions.
Once the environment changes, well-trained detectors may suffer performance degradation due to the existence of \emph{data distribution shifts} between training images and test images. As shown in Figure~\ref{Absfig}, the original detector performs well in the ideal (sunny) scenario while failing to maintain stable performance in the Snow and Fog scenario due to the out-of-distribution (OOD) issue. 
Considering the widespread application of 3D detection, significant performance degradation on OOD test data may result in traffic accidents and serious safety risks. Thus, it is essential to address the OOD generalization problem of vision-centric 3D detection.

\begin{figure*}[t] 
  \centering
  \includegraphics[width=\linewidth]{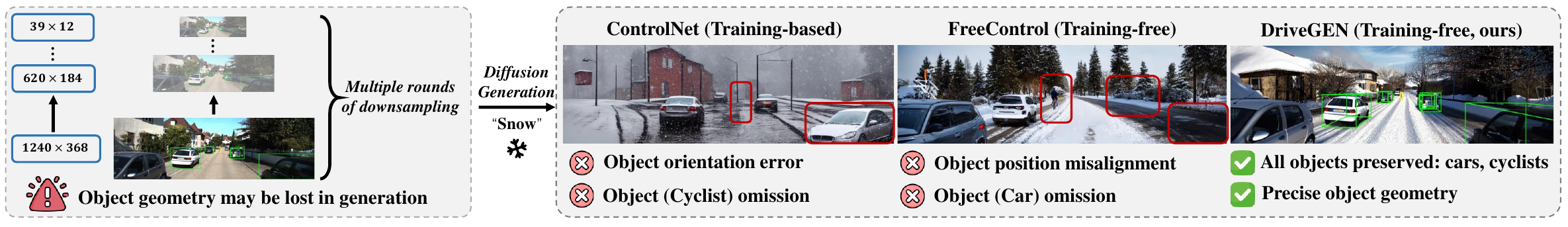} 
   \caption{
   % Illustration of object preservation challenges for 3D object detection during diffusion on KITTI.
    Illustration of the importance of object preservation in 3d detection based on KITTI.
   ControlNet~\cite{zhang2023adding} suffers object orientation errors and omissions even if accurate segmentation masks~\cite{ravi2024sam} and rich prompts~\cite{chen2024internvl} are provided, showing that training-based methods may struggle with spatial control with limited training data. 
   Additionally, Freecontrol~\cite{mo2024freecontrol} relies on coarse low-resolution features to capture semantic structures that may encounter potential object geometry loss, resulting in object position misalignment and omission issues.
   % The training-based controllable T2I approach, \ie ControlNet, suffers the object orientation error and the object omission even if the accurate segmentation masks~\cite{ravi2024sam} and the rich prompts~\cite{chen2024internvl} are provided for training,
   }
   \label{fig:our_motivation}
   % \vspace{-0.1in}
\end{figure*}

Recently, two attempts sought to solve the distribution shifts in AD by test-time model adaptation~\cite{lin2025monotta} or restoring OOD scenarios to ideal conditions with a weather-adaptive diffusion model~\cite{oh2025monowad}.
% adopting a weather-adaptive diffusion model to restore OOD scenarios to ideal conditions~\cite{oh2025monowad}. 
However, they require additional consumption during the test phase, leading to additional energy costs while the driving range is crucial in AD.
To save test-time computation costs, we explore controllable Text-to-Image (T2I) diffusion generation for training data augmentation, thereby improving the robustness of 3D detectors. 
Prior training-based controllable T2I approaches such as ControlNet~\cite{zhang2023adding} and T2i-adapter~\cite{mou2024t2i} offer users fine-grained spatial control but rely on substantial training data to train auxiliary modules, making them impractical for handling OOD scenarios in AD.
As for training-free methods, they~\cite{tumanyan2023plug,mo2024freecontrol} 
leverage low-resolution self-attention features to capture image structures for diffusion model guidance. However, these features are relatively coarse for preserving object geometric characteristics of different classes in 3D object detection. As shown in Figure~\ref{fig:our_motivation}, the object geometry may be lost after multiple rounds of downsampling (32x), leading to severe orientation error, position misalignment and omissions, particularly for tiny objects.
% for preserving geometric characteristics of multiple objects from different classes in 3D object detection
% struggling to preserve geometric characteristics of multiple objects from different classes in 3D object detection and thus leading to object orientation error and position misalignment.
% In addition, object geometry may also be lost during multiple rounds of downsampling in diffusion, leading to object omission, particularly for tiny objects (e.g., the \emph{cyclist} in Figure 2).
Therefore, an eligible training-free controllable T2I diffusion method for training data enhancement in 3D detection should preserve all objects with precise 3D geometry.

To address these challenges, we propose a method namely \textbf{DriveGEN} for Generalized and Robust 3D Detection in \textbf{Driv}ing via Training-Free Controllable Text-to-Image Diffusion \textbf{GEN}eration.
DriveGEN enhances the model robustness within various OOD scenarios generation, consisting of two stages: 
1) Self-Prototype Extraction. To capture accurate object geometric characteristics, we first extract principal components of self-attention features~\cite{tumanyan2023plug,mo2024freecontrol} to obtain coarse semantic-aware features. Then, we adopt layouts (\ie bounding boxes) to re-weight features in object regions via a peak function, termed self-prototypes.
2) Prototype-Guided Diffusion. To preserve objects with precise 3D geometry, we perform semantic-aware feature alignment with self-prototypes during the denoising phase.
However, these semantic-aware features are relatively coarse to represent tiny objects as shown in Figure~\ref{fig:our_motivation}. Therefore, we devise a shallow feature alignment strategy to constrain the diffusion process and retrain tiny objects.

\textbf{Contributions:} 1) To the best of our knowledge, we are the first to explore training-free controllable T2I diffusion generation to enhance the model robustness of vision-centric 3D detection. Even with a single augmentation, \ournet~improves the 3D detector~\cite{zhang2021objects} by an average of {7.6 mAP} across 13 OOD scenarios.
% 2) DriveGEN presents the first training-free solution that supports OOD scenario augmentation while preserving all objects with precise 3D geometry, thus improving the practical applicability of diffusion models in autonomous driving.
% 2) DriveGEN extracts self-prototypes to capture object geometries and  
% guides the diffusion process by aligning image features with self-prototypes, thus preserving precise 3D geometry for objects.
2) With self-prototype extraction and prototype-guided diffusion stages, \ournet~enables OOD scenario augmentation while preserving precise 3D geometry for all objects, enhancing the applicability of diffusion models in autonomous driving.
3) Extensive experiments show DriveGEN significantly improves existing monocular 3D detectors across 13 OOD corruption on KITTI and multi-view 3D detectors on the real scenarios (Night, Rainy) of nuScenes, demonstrating the effectiveness of \ournet~in boosting model generalizability.

% Free-DriveGEN brings sufficient performance improvement to 3D detectors in 12 unknown distributions  with only one augmentation scenario.
% in both known distributions fand unknown distributions.
%-------------------------------------------------------------------------------------------
%-------------------------------------------------------------------------------------------

\section{Related Work}
\label{sec:rw}

We first review the literature on explorations of model robustness in autonomous driving, and then discuss controllable T2I diffusion generation methods. More discussions on vision-centric 3D object detection are in Appendix~\ref{sec:supp_related}.

\noindent
\textbf{Model Robustness in Autonomous Driving.}
The reliability of vision-centric autonomous driving systems depends on the robustness of perception models, especially under challenging scenarios like adverse weather, varied lighting, and complex urban environments~\cite{xie2025benchmarking}.
To improve robustness, existing methods employ diffusion models to generate synthetic driving scenes under various conditions for training data augmentation.
Specifically, BEVGen~\cite{swerdlow2024street} and BEVControl~\cite{yang2023bevcontrol} employ bird’s-eye view layouts to create flexible scene configurations.
Innovative frameworks like Panacea~\cite{sun2022panacea} and MagicDrive~\cite{hong2021magicdrive} extend this controllability to multi-view and 3D scenes, while models such as DriveDreamer~\cite{zhao2022drivedreamer} and DrivingDiffusion~\cite{kim2021drivingdiffusion} focus on generating realistic videos for model testing. Recently, RoboFusion~\cite{song2024robofusion} has sought to tackle OOD noise scenarios by leveraging visual foundation models like SAM.

\noindent
\textbf{Controllable T2I Diffusion.}
Controllable text-to-image (T2I) diffusion models~\cite{rombach2022high,zheng2024memo,zhang2023hipa} are increasingly applied in real-world scenarios, where text prompts or scene descriptions guide the creation of synthetic data. By leveraging pre-trained models like Stable Diffusion~\cite{rombach2022high} or other large-scale T2I models~\cite{ramesh2022hierarchical,saharia2022photorealistic}, existing methods can generate high-fidelity and diverse images based on textual input, allowing for controlled variations~\cite{zhang2023expanding,chen2024human}. 
Techniques such as Uni-ControlNet~\cite{zhao2023uni}, UniControl~\cite{qin2023unicontrol} and Layoutdiffusion~\cite{zheng2023layoutdiffusion} further enhance control by integrating multimodal inputs, creating contextually accurate and condition-specific scenes.
Recent works also explore the utility of cross-attention mechanisms and latent embeddings to align generated images closely with text descriptions, supporting the generation for perception tasks like semantic segmentation and depth estimation~\cite{huang2022diffumask,li2021vpd}. 
However, these methods require a large amount of training data to train auxiliary modules, making them impractical for handling OOD scenarios in AD for the limited data scale.

Another alternative choice is training-free controllable T2I diffusion, PnP~\cite{tumanyan2023plug} directly injects the self-attention features of condition images to guide the diffusion process while Freecontrol~\cite{mo2024freecontrol} utilizes PCA to extract principle components of self-attention features for semantic structures extraction.
However, existing approaches solely rely on the relatively coarse self-attention features, which often results in object region omissions and misalignment as we mentioned before.
To this end, we propose our {DriveGEN} to preserve all objects with precise 3D geometry.
% Consequently, while these methods enable T2I diffusion to better match real-world driving complexities, there remains a need for greater control precision, semantic consistency, and computational efficiency to support the demands of autonomous driving applications.
% This approach facilitates the generation of specific, safety-critical driving scenarios by conditioning the diffusion process on customizable parameters.

% 1.  SimGen: Simulator-conditioned Driving Scene Generation arxiv + 不开源
% 这个方法更倾向于通用数据生成，不存在严格意义上的training data（或者说能收到的所有数据，都算是他的training data）
% 2.  Improving End-To-End Autonomous Driving with Synthetic Data from Latent Diffusion Models CVPR 24 workshop + 不开源（2D分割）
% 具体做法Both images and masks are resized to 512 × 512 以克服diffusion model的分辨率问题 --》 我们也需要好好想想这里要怎么处理，是不是同样resize，还是取patch呢？
% 3.  DatasetDM: Synthesizing Data with Perception Annotations Using Diffusion Models NIPS23 + 开源
% 这个框架偏向于分割，做从0-》1的训练数据生成，我想要的是类似于方法2但是能用于3D的生成框架设计
% 4.  DrivingDiffusion: Layout-Guided multi-view driving scene video generation with latent diffusion model Arxiv + 不开源
% 这个是根据Multi-Agent Trajectories & Road Structure 去生成Multi-View Videos/frames的框架，跟我们的想法还是一样的，我们不做0-》1的训练数据生成，而是要extend那些corruption cases
% 5.  DriveDiTFit: Fine-tuning Diffusion Transformers for Autonomous Driving Arxiv+不开源
% 这篇跟2一样，是我们到时候需要参考的方法之一；它做的是针对2D检测的corruption data generation

\section{Preliminary}
\label{sec:prel}

% \noindent
% \textbf{Diffusion Generation.}

Pre-trained diffusion models~\cite{ho2020denoising,dhariwal2021diffusion,rombach2022high} are probabilistic generative models, which consists of two complementary stochastic processes:
1) In the \emph{forward} process, Gaussian noise is incrementally added to a clean image $\mathbf{x}_0$.
2) As for the \emph{backward} process, it conducts the iterative process of denoising the initial Gaussian noise image $\mathbf{z}_{N_t}$, where a cleaner image is given at each step. To be specific, this process leverages a denoising network $\epsilon_{\theta}(\mathbf{z}_t, t)$ to estimate the added noise in the forward process. Subsequently, for $\mathbf{z}_t$ at each step, the network removes the estimated noise perturbation to achieve a cleaner $\mathbf{z}_{t-1}$.

For T2I diffusion models, previous method~\cite{songscore} has shown that $\epsilon_{\theta}$ can approximate the score function for the marginal distributions $p_t$, \ie $- \sigma_t \nabla_{\mathbf{z}_t} \log p_t(\mathbf{z}_t | \mathbf{c})$ where $\sigma_t$ denotes a noise schedule and $\mathbf{c}$ is a text prompt.
To incorporate auxiliary information $y$ into the diffusion sampling process, existing methods~\cite{dhariwal2021diffusion,epstein2023diffusion,mo2024freecontrol} guide the diffusion process by adding a time-dependent energy function termed $g(\mathbf{z}_t; t, y)$ with a strength coefficient $s$: 
\vspace{-0.1in}

$$
\hat{\epsilon}_{\theta}(\mathbf{z}_t; t,c)\small{=}{\epsilon}_{\theta}(\mathbf{z}_t; t,c)-s\cdot g(\mathbf{z}_t; t, y),
$$

where $g$ can be defined via bounding boxes~\cite{xie2023boxdiff}.
In this paper, we adopt a standard choice $\epsilon_{\theta}$ following existing methods~\cite{tumanyan2023plug, mo2024freecontrol}, \ie U-Net architecture~\cite{ronneberger2015u} with multi-level self-attention and cross-attention layers~\cite{vaswani2017attention}.

\section{Training-Free Controllable T2I Diffusion}
\label{sec:method}

%-------------------------------------------------------------------------------------------
\begin{figure*}[t] 
  \centering
  \includegraphics[width=0.95\linewidth]{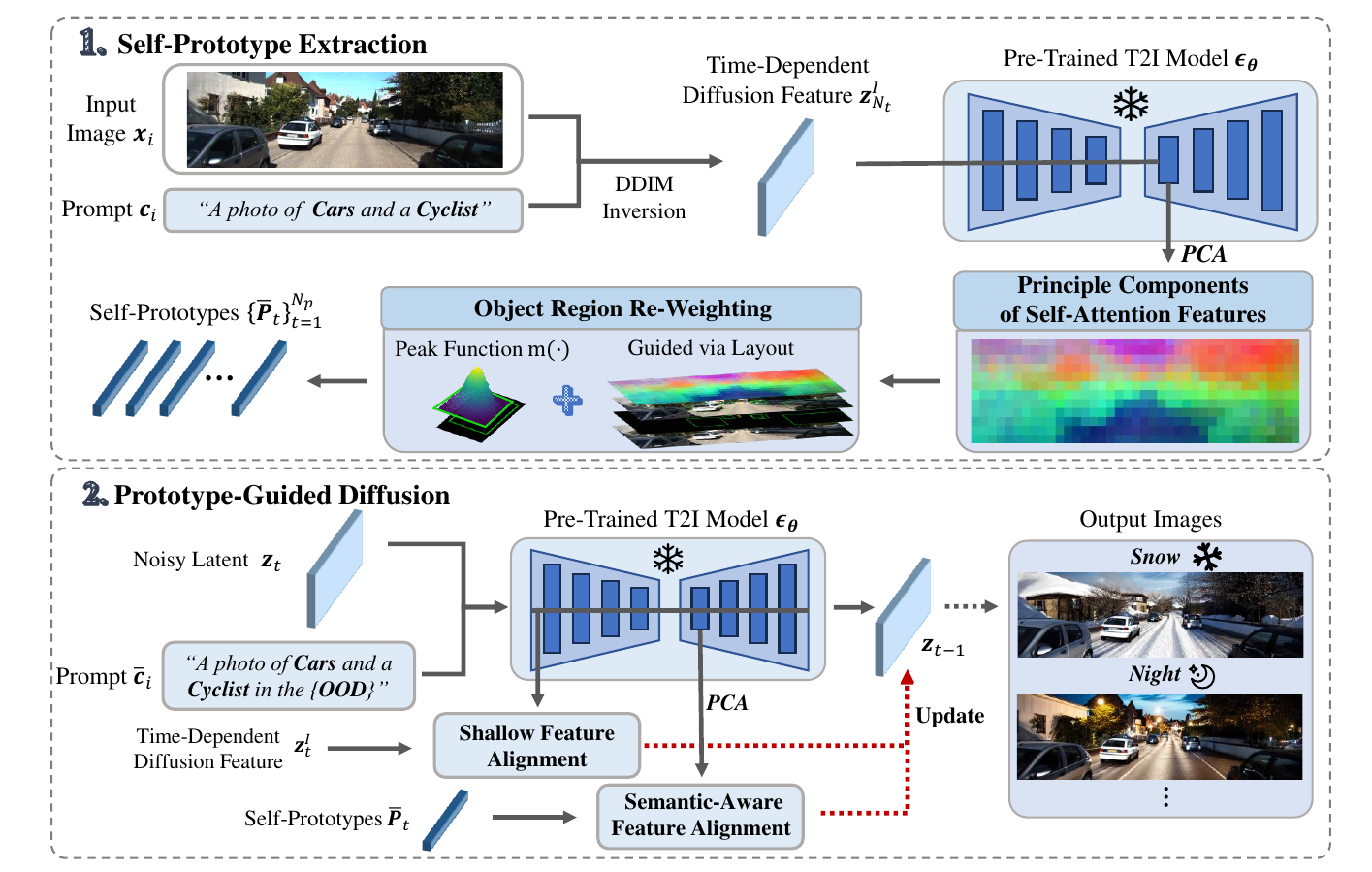} 
   \caption{
   An overview of our \ournet
   % . It aims to improve the robustness of 3D detectors by training data enhancement with the training-free diffusion process
   , consisting of two stages:
   1) The Self-Prototype Extraction stage is devised to extract accurate semantic structures of multiple objects. To capture precise locations, we achieve fine-grained self-prototypes via 
    leveraging layouts and the peak function to re-weight object regions rather than directly using coarse self-attention features.
   2) The Prototype-Guided Diffusion stage conducts semantic-aware feature alignment for semantic structure matching and shallow feature alignment for tiny object preservation.
   }
   \label{fig:our_method}
   % \vspace{-0.05in}
\end{figure*}
%-------------------------------------------------------------------------------------------

\noindent
% \textbf{Problem statement.}
\subsection{Problem Statement}
Without loss of generality, we denote the  well-trained 3D detector as $f_{\Theta_{d}}(\cdot)$ which is obtained by training on labeled training images $\mathcal{D}\small{=}\{(\mathbf{x}_i, \mathbf{y}_i)\}_{i=1}^{N}$.
The training images follow the original training distribution $P\left(\mathbf{x}\right)$ (\ie~$\mathbf{x} \sim P\left(\mathbf{x}\right)$).
Here, $\Theta_{d}$ represents the parameters of the detector and $N$ is the total number of training images.
After training, the model is applied to unlabeled test images $\mathcal{D}_t\small{=}\{\mathbf{x}^t\}$.
% $\mathcal{D}_t\small{=}\{\mathbf{x}_i^t\}_{i=1}^{N_t}$  where $N_{t}$ is the total number of test images.
Unfortunately, the well-trained model often encounters Out-of-distribution (OOD) test scenarios in autonomous driving (AD) due to prevalent natural corruptions, namely distribution shifts,
\ie~$\mathbf{x}^t \sim Q\left(\mathbf{x}\right)$ and $P\left(\mathbf{x}\right) \neq Q\left(\mathbf{x}\right)$.

% To solve this problem, MonoTTA~\cite{lin2025monotta} conducts online test-time model adaptation to alleviate data distribution shifts in OOD scenarios while requiring one round of backward computation for each test image. In addition, MonoWAD~\cite{oh2025monowad} solves this issue by utilizing a weather-adaptive diffusion model to restore each test image to the original training distribution. However, these methods are relatively time-intensive for the additional computation costs during the test phrase, which probably reduce the cruising range of vehicles.
% To summarize, existing approaches are unable to enhance the robustness of 3D object detectors solely through the training stage, without any additional computation consumption at the test stage.

As previously discussed, existing methods~\cite{lin2025monotta,oh2025monowad} are relatively time-consuming due to their additional computation costs at test time. 
Hence, we investigate controllable Text-to-Image (T2I) diffusion generation for training data augmentation, thereby improving the robustness of 3D detectors. Particularly, it is crucial to preserve object information because the annotations would be reused during detector training.
However, training-based approaches~\cite{zhang2023adding,mou2024t2i} require substantial training data to train auxiliary modules for spatial control, which is often unsatisfying in AD.
Besides, training-free methods~\cite{tumanyan2023plug,mo2024freecontrol} mainly consider the relatively coarse self-attention features, making it challenging to preserve relatively tiny objects.

\subsection{Overall Scheme}
After examining the characteristics and challenges of robust 3D object detection, we introduce \textbf{DriveGEN}, a Training-Free Controllable Text-to-Image Diffusion Generation method devised to enhance training data and improve the robustness of vision-centric 3D detectors. 
As illustrated in Figure~\ref{fig:our_method}, \ournet~consists of two stages: 1) Self-Prototype Extraction and 2) Prototype-Guided Diffusion. We briefly introduce each stage below.

First, we develop the self-prototype extraction stage (c.f. Section~\ref{sec:stage1}) to capture the geometric characteristics of objects accurately. Existing methods apply principal component analysis (PCA) on the self-attention features of the first decoder layer of the diffusion model $\epsilon_{\theta}$ to capture semantic structures of condition images~\cite{tumanyan2023plug,mo2024freecontrol}.
However, we empirically find that solely relying on cross-attention relevance between self-attention features and object prompts for region selection may result in object preservation errors (c.f. Figure~\ref{fig:our_motivation}).
To address this, \ournet~leverages layouts to re-weight object regions in the principal components of self-attention features via a peak function and then achieves the self-prototypes which help the diffusion model focus on the accurate object regions during generation.

% Given the input image, we generate the text prompt via its annotation and then undergo DDIM inversion~\cite{song2020denoising} to achieve the initial time-dependent diffusion feature.
% With the pre-trained T2I diffusion model $\epsilon_{\theta}$, we utilize principal component analysis (PCA) to extract principal components of the self-attention features of the first decoder layer which is able to represent semantic structures of condition images~\cite{tumanyan2023plug,mo2024freecontrol}.
% However, we empirically find that they are solely on the relevance (\ie cross-attention layers) between self-attention features and object prompts for the object region selection, which often leads to object preservation errors as shown in Figure~\ref{fig:our_motivation}.
% principal components are too coarse to represent the geometry of multiple objects, particularly in 3D object detection as shown in Figure~\ref{fig:our_method}.
% This limitation 
% To address this, DrivenGEN leverages layouts to re-weight object regions in the principal components of self-attention features via a peak function and then achieves the self-prototypes which help the diffusion model focus on the accurate object regions during generation.
% , thereby ensuring the accurate extraction of self-prototypes $\mathbf{P}_T$ for object preservation.
% those regions of interested.
% Subsequently, Free-DrivenGEN achieves the self-prototypes of different self-attention modules which represents accurate object geometry.

% \vspace{-0.1in}
Second, we introduce the prototype-guided diffusion stage (c.f. Section~\ref{sec:stage2}) to preserve all objects during the generation process.
On the one hand, we conduct semantic-aware feature alignment between self-attention features and self-prototypes to capture accurate object semantic structures.
On the other hand, \ournet~performs shallow feature alignment to further preserve object information by constraining the noisy latent at the shallow level, compensating for the gaps in semantic-aware feature alignment regarding tiny object preservation.
% On the one hand, we conduct semantic-aware feature alignment based on the principal components of self-attention features and our self-prototypes to capture accurate object semantic structures. 
% On the other hand, low-resolution self-attention features present a challenge for retaining detailed geometric characteristics about tiny objects as demonstrated in the left side of Figure~\ref{fig:our_motivation}.
% Hence, DrivenGEN further performs shallow feature alignment to constrain the noisy latent at the shallow level, compensating for the gaps in semantic-aware feature alignment regarding tiny object preservation.
% we conduct self-attention feature (\ie the principal components) alignment for object preservation during the denoising phase with our self-prototypes.
% However, as shown in Figure~\ref{fig:our_motivation}, the low spatial resolution of principal components makes it challenging to fully preserve information on tiny objects at this level due to multiple rounds of downsampling.
% To this end, Free-DrivenGEN conducts shallow feature alignment before downsampling, which constraints the noisy latent to retain fine-grained object information and compensates the absence of self-attention feature alignment on tiny object preservation.
The Pseudo-code of \ournet~is summarized in Algorithm~\ref{al:training}.

\subsection{Self-Prototype Extraction}
\label{sec:stage1}
To accurately capture the geometric characteristics of multiple objects, we propose a self-prototype extraction strategy that consists of two components: 
1) Image structure extraction and 2) Object region re-weighting.

\noindent
\textbf{Image Structure Extraction.} 
To achieve zero-shot spatial control, capturing the semantic structures of input images is essential. Prior works~\cite{tumanyan2023plug,mo2024freecontrol} demonstrate that self-attention features (\eg keys) of the T2I diffusion model $\epsilon_{\theta}$ are semantic-aware~\cite{rombach2022high}.
Thus, given the input image $\mathbf{x}_i$, we first generate the text prompt $\mathbf{c}_i$ via a simple template \emph{`A photo of \{object\}, ..., and \{object\}'.}
Then, we leverage DDIM Inversion~\cite{song2020denoising} to achieve a series of time-dependent diffusion features $\{\mathbf{z}^I_{t} \in \mathbb{R}^{N_c \times H \times W}\}^{N_t}_{t=1}$ where $N_t$ denotes the number of DDIM inversion and sampling steps, $N_c$ represents the number of feature channels, and $H,W$ indicate the height and width, respectively.
% utilize DDIM Inversion~\cite{song2020denoising} to achieve a series of time-dependent diffusion features $\{\mathbf{z}^I_{t} \in \mathbb{R}^{N_c \times H \times W}\}^{N_t}_{t=1}$ where $N_t$ denotes the number of DDIM inversion and sampling steps, $N_c$ represents the number of feature channels, $H,W$ indicate the height and width. 
% Note $\mathbf{c}_i$ is generated via a simple template \emph{`A photo of \{object\}, ..., and \{object\}'.}
During sampling, we apply PCA on self-attention features to extract principal components $\{\mathbf{P}_t\}^{N_p}_{t=1}$, each $\mathbf{P}_t\in\mathbb{R}^{N_b \times H \times W}$ and $\mathbf{P}_t=\{ \mathbf{p}^1_t, \dots, \mathbf{p}^{N_b}_t \}$. Here $N_b$ denotes the number of principal components and we store $\mathbf{P}_t$ of the initial $N_p$ steps.

% To this end, given the input image $\mathbf{x}_0$ and the text prompt $\mathbf{c}$, we first utilize DDIM Inversion~\cite{song2020denoising} to achieve a series of time-dependent diffusion features $\{\mathbf{x}_i^I\}_{i=1}^{T}$. With the final inversion feature $\mathbf{x}_T^I$, we obtain the self-attention features through the pretrained T2I diffusion model $\epsilon_{\theta}$. To dig out the semantic structure, we perform PCA to extract principal components $\mathbf{P}_T$ of self-attention features.
\noindent
\textbf{Object Region Re-weighting.}
Existing methods~\cite{ge2023expressive,mo2024freecontrol} select foreground regions by leveraging cross-attention maps between specific words (\ie object names) and image features to derive a concept region mask $\mathbf{M}$.
However, relying solely on such attention maps may result in omissions and misalignment of object regions, as illustrated in Figure~\ref{fig:our_motivation}.
To address this, we introduce a peak function $m(\cdot)$ to re-weight $\mathbf{M}$ for each object region based on layouts $\mathbf{L}$ after the coordinate transformation through downsampling.
To be specific, given a point $(p, q)$ from $\mathbf{L}_i$ of $\mathbf{x}_i$, the corresponding object center is defined as $(\hat{p}, \hat{q})$ and then we can calculate $m(p, q)$ by:
% $\mathbf{x}_i^{(p, q)}$ selected from $\mathbf{L}_i$ with its object center $\mathbf{x}_i^{(\hat{p}, \hat{q})}$, we can calculate $m(\mathbf{x}_i^{(p, q)})$:
% \vspace{-0.05in}
\begin{equation}
\scalebox{1}{$
\begin{aligned}
\label{eqn:peak}
 m(p, q) = \exp \left( -\frac{(p - \hat{p})^2 + (q - \hat{q})^2}{2\sigma^2} \right)
\end{aligned}
$}
\end{equation}
where $\sigma \in (0, 1)$ controls the range of the influence area. Then, we update the concept token mask $\mathbf{M}$ by:
\begin{equation}
\label{eqn:mask}
\hat{\mathbf{M}}(p, q) = \mathbf{M}(p, q) + m(p,q)\cdot\mathbf{L}_i(p,q).
\end{equation}
% Here, $\mathbb{I}(\cdot)$ is the indicate function. 
The object regions in $\mathbf{L}_i$ are marked as $1$ while all other regions as 
$0$.
Eventually, we obtain the self-prototypes by $\{\hat{\mathbf{P}}_t\}^{N_p}_{t=1}=\{\mathbf{P}_t\}^{N_p}_{t=1}\big|_{\mathbf{\hat{M}}}$ which guide the diffusion model to localize precise locations of objects and combine $\mathbf{\hat{M}}$ for accurate object region selection.

\begin{algorithm}[t]
    \footnotesize
    \caption{The pipeline of the proposed \ournet}\label{al:training}
    \begin{algorithmic}[1]
        \REQUIRE Training data $\{(\mathbf{x}_i^s, \textbf{y}_i^s)\}_{i=1}^{N}$; Pretrained T2I diffusion model $\epsilon_{\theta}$; Hyper-parameters $s$, $\sigma$, $\tau$; The OOD scenario.
    \\
    \#\# Stage 1: Self-Prototype Extraction \#\#
    \FOR {each training image $\mathbf{}{x}_i$} 
        \STATE Get the text prompt $c_i$ and layout $\mathbf{L}_i$ based on $\mathbf{y}_i$; 
        \STATE Undergo DDIM Inversion to get $\mathbf{P}_t$;
        \STATE Get updated $\hat{\mathbf{M}}$ via Eqn.~(\ref{eqn:peak}), Eqn.~(\ref{eqn:mask}) for self-prototypes  $\hat{\mathbf{P}}_t$.
    \ENDFOR
    \\
    % \% Stage 2: Prototype-Guided Diffusion
    \#\# Stage 2: Prototype-Guided Diffusion \#\#
    \FOR {each training image $\mathbf{x}_i$}
        \STATE Update $\mathbf{c}_i$ with specific OOD type;
        \FOR {diffusion step $t = 1 \to N_t$}
            \IF{$t \leq N_p$}
                \STATE Calculate $g_{sa}$, $g_{sl}$ based on Eqn.~(\ref{eqn:sa_l2}), Eqn.~(\ref{eqn:sl});
                \STATE Get the gradient based on Eqn.(\ref{eqn:overall});
            \ENDIF
            \STATE Update the noisy latent $\mathbf{z}_t \to \mathbf{z}_{t-1}$;
        \ENDFOR
    \ENDFOR
    
    \RETURN Output images for all $\mathbf{x}_i$ in the OOD scenario.
     \end{algorithmic}
\end{algorithm}

\subsection{Prototype-Guided Diffusion}
\label{sec:stage2}
To preserve all objects with precise 3D geometry during diffusion, we leverage the self-prototypes $\{\hat{\mathbf{P}}_t\}^{N_p}_{t=1}$ and time-dependent diffusion features $\{\mathbf{z}_t^I\}_{t=1}^{N_t}$ for feature alignment, including: 1) semantic-aware feature alignment and 2) shallow feature alignment.
% With the self-prototypes, we are able to conduct feature alignment to guide the diffusion process for multiple object preservation, including 1) self-attention feature alignment and 2) shallow feature alignment.

\noindent
\textbf{Semantic-aware Feature Alignment.}
As mentioned before, the self-attention features of the first decoder layer can represent the semantic structures of input images.
Thus, we first initialize the noisy latent $\mathbf{z}_{N_t}$ via the $\mathbf{z}^I_{N_t}$. For the step $t\leq N_p$, we perform PCA to obtain the principal components $\mathbf{P}_t$ of the self-attention features. 
Then, we adopt a pre-defined threshold $\tau$ following~\cite{ge2023expressive} to select informative regions by $\hat{\mathbf{M}} = \mathbb{I}(\hat{\mathbf{M}} > \tau)$.
% $$
% \hat{\mathbf{M}} = \mathbb{I}(\hat{\mathbf{M}} > \tau),
% $$
% where $\mathbb{I}(\cdot)$ is the indicate function.
Specifically, we calculate the semantic-aware feature alignment loss with $\mathbf{P}_t$ and $\hat{\mathbf{P}}_t$:
\begin{equation}
\label{eqn:sa_l2}
g_{sa} = \|\hat{\mathbf{M}} \odot (\mathbf{P}_t - \hat{\mathbf{P}}_t)\|_2^2,
\end{equation}
where $\odot$ represents element-wise multiplication, applying the mask $\hat{\mathbf{M}}$ to the  difference between $\mathbf{P}_t$ and $\hat{\mathbf{P}}_t$.

\noindent
\textbf{Shallow Feature Alignment.}
% Although the alignment of self-attention features guides the diffusion model to capture object semantic structures,
As shown in Figure~\ref{fig:our_motivation}, we empirically find that principal components lack sufficient details to represent objects, especially tiny objects. For instance, given a bounding box of a cyclist with a height of 20 pixels and a width of 5 pixels, this object is unable to occupy even one element within principal components after multiple rounds of downsampling (e.g., 32x).

Therefore, \ournet~introduces shallow feature alignment before the forward process of diffusion, which aims to preserve fine-grained object details by constraining the noisy latent.
Specifically, based on the layout $\mathbf{L}_i$ with $N_i$ objects, we calculate the shallow feature alignment loss by:
\begin{equation}
\label{eqn:sl}
% g_{sl} = \frac{1}{N_i}\|\mathbf{L}_i \odot (\mathbf{z}^I_{t} - \mathbf{z}_{t})\|^2_2,
g_{sl} {=} \frac{1}{N_i} {\sum_{p,q}} \|\mathbf{L}_i(p,q) \odot (\mathbf{z}^I_{t}(p,q) - \mathbf{z}_{t}(p,q))\|^2_2,
\end{equation}

Overall, the total scheme of \ournet~is as follows:
\begin{equation}
\label{eqn:overall}
\hat{\epsilon}_t={(1+s)\epsilon}_{\theta}(\mathbf{z}_t; t,c)-s\epsilon_{\theta}(\mathbf{z}_t; t, \emptyset)+ g_{sa} +  g_{sl}.
\end{equation}
Note that $\emptyset$ denotes null and \ournet~is a classifier-free guidance~\cite{ho2022classifier} process with hyper-parameters $s$, $\sigma$ and $\tau$.

% \vspace{-0.15in}
\begin{table*}[t]
\setlength\tabcolsep{9pt}
\renewcommand\arraystretch{0.9}
    \begin{center}
    \caption{\label{tab:kitti-c} Comparison on the {KITTI-C} dataset, severity \textbf{level 1} regarding {Mean $AP_{3D|R_{40}}$}. The \textbf{bold} number indicates the best result. 
    % \textcolor{blue}{Blue} cells show scenarios with similar training augmentation data for the baseline while other cells represent unseen scenarios.
    }  
    \vspace{-0.05in}
    \scalebox{0.56}{
         \begin{tabular}{lc|ccc|ccc|cccc|ccc|c}
         \toprule
         \multicolumn{15}{c}{\textbf{\textbf{Car}, IoU @ 0.7, 0.5, 0.5}} \\
         \midrule
         \multirow{2}{*}{Method}  &
         \multirow{2}{*}{\shortstack{Training-free\\diffusion}} &
         \multicolumn{3}{c|}{Noise} & 
         \multicolumn{3}{c|}{Blur} & 
         \multicolumn{4}{c|}{Weather} & 
         \multicolumn{3}{c|}{Digital} & 
         \multirow{2}{*}{Avg.} \\
        \cmidrule(lr){3-5} \cmidrule(lr){6-8} \cmidrule(lr){9-12} \cmidrule(lr){13-15} 
        & & Gauss. & Shot & Impul. & Defoc. & Glass & Motion & Snow & Frost & Fog & Brit. & Contr. & Pixel & Sat. \\
         \midrule
        Monoflex~\cite{zhang2021objects} &  &  13.06 &  20.91 &  14.09 &  20.17 &  28.59 &  30.34 &  33.64 &  30.31 &  19.58 &  45.22 &  20.01 &  29.07 &  38.85 & 26.45 \\
        \midrule
        
        ~$\bullet~$ Color Jitter (Traditional aug.) & & 9.55 & 15.81 & 11.90 & 22.67 & 25.38 & 30.12 & 34.08 & 30.00 & 19.29 & 42.10 & 19.93 & 17.17 & 36.48 & 24.19  \\
        ~$\bullet~$ Brightness (Traditional aug.) & & 11.44 & 19.42 & 12.73 & 11.18 & 18.95 & 22.03 & 26.64 & 21.70 & 13.04 & 39.61 & 13.08 & 21.11 & 29.73 & 20.05 \\
        
        \midrule
        ~$\bullet~$ ControlNet (Only Snow aug.) & \xmark & 0.32 & 1.18 & 1.28 & 4.65 & 11.32 & 17.04 & {22.84} & 19.72 & 9.53 & 34.79 & 8.73 & 1.25 & 16.88 & 11.50  \\  
        ~$\bullet~$ ControlNet (3 scenarios aug.) & \xmark & 0.70 & 1.14 & 0.29 & 0.35 & 0.37 & 0.83 &  4.07 & 3.05 &  1.20 & 9.43 & 0.94 & 0.48 & 3.20 & 2.00  \\  
        ~$\bullet~$ ControlNet (6 scenarios aug.) & \xmark & 0.00 & 0.00 & 0.00 &  0.00 & 0.00 & 1.94 &  0.00 & 0.00 &  0.00 & 0.00 & 0.00 & 0.00 & 0.00 & 0.15 \\  
         \midrule
        ~$\bullet~$ Freecontrol (Only Snow aug.) & \cmark & 20.39 & 28.05 & 22.27 & 12.52 & 21.67 & 21.09 & 27.97 & 17.91 & 10.07 & 35.99 & 9.90 & 28.56 & 35.27 & 22.44 \\  
        ~$\bullet~$ Freecontrol (3 scenarios aug.) & \cmark & 15.43 & 21.25 & 14.04 & 15.33 & 19.22 & 16.53 & 22.01 & 17.41 & 15.18 & 24.66 & 16.66 & 25.39 & 31.44 & 19.58 \\  
        ~$\bullet~$ Freecontrol (6 scenarios aug.) & \cmark & 12.85 & 18.71 & 16.31 & 12.06 & 16.31 & 12.86 & 17.64 & 16.18 & 14.23 & 24.52 & 15.64 & 22.56 & 24.28 & 17.24 \\  
         \midrule
        ~$\bullet~$ DriveGEN (Only Snow aug.) & \cmark & 16.48 & 26.72 & 24.98 & 31.17 & 35.55 & 38.13 &  \textbf{41.39} & {38.61} & 27.85 & \textbf{49.76} & 29.28 & 38.88 & \textbf{44.12} & 34.07  \\  
        ~$\bullet~$ DriveGEN (3 scenarios aug.) & \cmark & \textbf{25.63} & \textbf{37.04} & \textbf{29.13} & {34.13} & {39.15} & 36.81 & 38.58 & 37.98 & 33.93 & 45.39 & 34.66 & 39.36 & 43.81 & 36.58  \\  
        ~$\bullet~$ DriveGEN (6 scenarios aug.) & \cmark & 24.77 & 33.79 & 28.27 & \textbf{36.92} &\textbf{40.33}& \textbf{40.45} & {40.60} & \textbf{40.56} & \textbf{38.10} & 44.83 & \textbf{39.28} & \textbf{41.81} & 44.05 & \textbf{37.98} \\ 
         \midrule
        \midrule
        MonoGround~\cite{qin2022monoground} &  & 13.05 & 21.77 & 18.87 & 20.79 & 30.74 & 32.02 & 34.43 & 27.02 & 14.15 & 46.21 & 14.63 & 33.41 & 35.60 & 26.36  \\
        \midrule
        
        ~$\bullet~$ Color Jitter (Traditional aug.) & & 12.88 & 24.31 & 18.95 & 23.07 & 30.44 & 31.42 & 35.94 & 30.43 & 19.89 & 44.66 & 20.61 & 29.75 & 36.65 & 26.36 \\
        ~$\bullet~$ Brightness (Traditional aug.) & & 14.02 & 23.52 & 20.14 & 23.95 & 31.78 & 28.79 & 35.08 & 31.87 & 18.87 & 42.94 & 17.75 & 25.55 & 37.18 & 27.03 \\
        
        \midrule
        ~$\bullet~$ ControlNet (Only Snow aug.) & \xmark &  1.76 & 3.23 & 4.63 & 5.20 & 12.95 & 14.11 & 17.70 & 11.58 & 3.04 & 35.21 & 2.98 & 7.29 & 13.98 & 10.28  \\  
        ~$\bullet~$ ControlNet (3 scenarios aug.) & \xmark & 0.00 & 0.00 & 0.24 & 1.42 & 1.68 & 1.61 & 4.90 & 5.40 & 0.57 & 17.90 & 1.12 & 7.79 & 6.82 & 3.80   \\  
        ~$\bullet~$ ControlNet (6 scenarios aug.) & \xmark & 0.00 & 0.00 & 0.00 & 1.68 & 1.26 & 0.35 & 1.13 & 0.52 & 0.44 & 4.08 & 0.38 & 2.22 & 1.77 & 1.06  \\  
         \midrule
        ~$\bullet~$ Freecontrol (Only Snow aug.) & \cmark & 11.75 & 21.89 & 15.76 & 17.70 & 21.45 & 21.69 & 32.08 & 20.60 & 13.57 & 36.05 & 14.03 & 26.75 & 38.35 & 22.43  \\  
        ~$\bullet~$ Freecontrol (3 scenarios aug.) & \cmark & 16.31 & 20.75 & 17.61 & 13.10 & 16.84 & 14.82 & 17.72 & 15.89 & 11.87 & 25.20 & 13.55 & 22.57 & 24.49 & 17.75 \\  
        ~$\bullet~$ Freecontrol (6 scenarios aug.) & \cmark & 15.20 & 22.59 & 15.35 & 22.00 & 21.18 & 18.95 & 17.69 & 14.85 & 14.82 & 24.02 & 16.97 & 22.99 & 26.12 & 19.44 \\  
         \midrule
        ~$\bullet~$ DriveGEN (Only Snow aug.) & \cmark & 17.07 & 26.78 & 23.78 & 32.89 & 37.52 & 39.06 & \textbf{40.61} & 34.91 & 25.29 & \textbf{46.21} & 27.12 & 38.25 & 44.45 & 33.38 \\  
        ~$\bullet~$ DriveGEN (3 scenarios aug.) & \cmark & 19.79 & 31.44 & 27.63 & 36.84 & 40.10 & 39.35 & 39.17 & 36.42 & 28.79 & 45.42 & 29.29 & 42.60 & 44.99 & 35.53  \\  
        ~$\bullet~$ DriveGEN (6 scenarios aug.) & \cmark & \textbf{23.84} & \textbf{32.59} & \textbf{30.34} & \textbf{38.57} & \textbf{41.20} & \textbf{40.19} & 38.16 & \textbf{38.40} & \textbf{32.53} & 43.95 & \textbf{34.80} & \textbf{44.10} & \textbf{45.13} & \textbf{37.21}  \\ 

        \midrule
        \midrule
        % \toprule
         \multicolumn{15}{c}{\textbf{\textbf{Pedestrian}, IoU @ 0.5, 0.25, 0.25}} \\
         \midrule
         \midrule
        %  \multirow{2}{*}{Method}  &
        %  \multirow{2}{*}{\shortstack{Training-free\\diffusion}} &
        %  \multicolumn{3}{c|}{Noise} & 
        %  \multicolumn{3}{c|}{Blur} & 
        %  \multicolumn{4}{c|}{Weather} & 
        %  \multicolumn{3}{c|}{Digital} & 
        %  \multirow{2}{*}{Avg.} \\
        % \cmidrule(lr){3-5} \cmidrule(lr){6-8} \cmidrule(lr){9-12} \cmidrule(lr){13-15} 
        % & & Gauss. & Shot & Impul. & Defoc. & Glass & Motion & Snow & Frost & Fog & Brit. & Contr. & Pixel & Sat. \\
        %  \midrule
          Monoflex~\cite{zhang2021objects} &  &   1.16 & 3.54 & 0.78 & 8.06 & 17.70 & 15.26 & 12.72 & 9.25 & 5.61 & 19.87 & 5.35 & 1.49 & 8.65 & 8.42 \\
        \midrule
        ~$\bullet~$ Color Jitter (Traditional aug.) & & 0.99 & 3.53 & 1.41 & 10.05 & 14.63 & 12.00 & 14.73 & 12.13 & 7.72 & 19.02 & 9.23 & 1.11 & 11.29 & 9.07  \\
        ~$\bullet~$ Brightness (Traditional aug.) & & 0.63 & 1.85 & 1.20 & 5.01 & 13.72 & 12.15 & 7.93 & 5.84 & 1.87 & 16.89 & 2.30 & 0.48 & 3.77 & 5.67\\
        \midrule          
          ~$\bullet~$ ControlNet (Only Snow aug.) & \xmark &   0.00 & 0.00 & 0.00 & 1.78 & 8.32 & 5.84 & 4.03 & 3.75 & 1.31 & 11.00 & 1.23 & 0.00 & 0.93 & 2.94  \\
           ~$\bullet~$ ControlNet (3 scenarios aug.) & \xmark &   0.00 & 0.00 & 0.00 & 1.15 & 3.23 & 1.14 & 1.64 & 1.35 & 0.83 & 4.06 & 2.01 & 0.00 & 0.00 & 1.19  \\
           ~$\bullet~$ ControlNet (6 scenarios aug.) & \xmark &   0.00 & 0.00 & 0.00 & 0.00 & 0.36 & 0.00 & 0.00 & 0.00 & 0.00 & 2.50 & 0.00 & 0.00 & 0.00 & 0.22  \\
           \midrule
           ~$\bullet~$ Freecontrol (Only Snow aug.) & \cmark & 3.62 & 5.17 & 3.21 & 3.23 & 5.16 & 5.94 & 4.68 & 3.52 & 4.28 & 7.70 & 4.22 & 8.51 & 7.70 & 5.15  \\
           ~$\bullet~$ Freecontrol (3 scenarios aug.) & \cmark & 2.26 & 3.54 & 3.04 & 3.27 & 4.97 & 4.25 & 7.42 & 4.41 & 7.27 & 10.91 & 6.18 & 6.83 & 7.43 & 5.52  \\
           ~$\bullet~$ Freecontrol (6 scenarios aug.) & \cmark & 5.99 & 9.10 & 9.08 & 4.80 & 7.36 & 5.84 & 7.22 & 7.58 & 7.10 & 10.81 & 9.40 & 9.90 & 10.24 & 8.03 \\
           \midrule
           ~$\bullet~$ DriveGEN (Only Snow aug.) & \cmark &  1.25 & 3.22 & 4.56 & 16.21 &\textbf{ 19.88} & \textbf{20.61} & \textbf{19.80} & 14.46 & 8.80 & \textbf{24.62} & 9.11 & 9.77 & 18.04 & 13.10  \\
           ~$\bullet~$ DriveGEN (3 scenarios aug.) & \cmark &  6.09 & \textbf{9.85} & \textbf{9.60} & 14.96 & 18.59 & 16.47 & 15.45 & \textbf{16.39} & \textbf{15.88} & 21.98 & \textbf{16.85} & \textbf{14.55} & \textbf{18.13} & 14.98  \\
           ~$\bullet~$ DriveGEN (6 scenarios aug.) & \cmark & \textbf{6.34} & 9.22 & 7.71 & \textbf{17.60} & 19.21 & 20.20 & 16.88 & 16.19 & 15.72 & 23.57 & 16.75 & 14.37 & 17.31 & \textbf{15.47}  \\
            \midrule
            \midrule
         MonoGround~\cite{qin2022monoground} &  & 2.67 & 3.25 & 5.76 & 17.57 & 18.91 & 17.71 & 12.96 & 9.35 & 4.37 & 24.15 & 5.89 & 3.27 & 7.16 & 10.23  \\
        \midrule
        
        ~$\bullet~$ Color Jitter (Traditional aug.) & & 2.44 & 3.24 & 4.11 & 15.37 & 18.46 & 16.50 & 15.45 & 12.38 & 9.14 & 24.71 & 9.82 & 2.07 & 7.81 & 10.89 \\
        ~$\bullet~$ Brightness (Traditional aug.) & & 2.79 & 4.10 & 7.61 & 14.51 & 14.13 & 14.52 & 12.12 & 12.66 & 5.55 & 20.80 & 5.36 & 2.36 & 10.31 & 9.75  \\
        
        \midrule         
          ~$\bullet~$ ControlNet (Only Snow aug.) & \xmark & 1.85 & 1.03 & 0.81 & 7.18 & 9.97 & 8.32 & 1.32 & 2.92 & 1.37 & 12.99 & 1.40 & 0.28 & 0.90 & 3.87  \\
           ~$\bullet~$ ControlNet (3 scenarios aug.) & \xmark & 0.00 & 0.00 & 0.00 & 3.60 & 3.89 & 1.96 & 1.14 & 1.03 & 1.50 & 4.21 & 1.25 & 0.26 & 0.77 & 1.51  \\
           ~$\bullet~$ ControlNet (6 scenarios aug.) & \xmark &  0.00 & 0.00 & 0.00 & 0.00 & 0.00 & 1.67 & 0.00 & 0.00 & 0.00 & 0.00 & 0.00 & 0.00 & 0.00 & 0.13   \\
           \midrule
           ~$\bullet~$ Freecontrol (Only Snow aug.) & \cmark & \textbf{10.04} & 13.09 & 11.70 & 14.91 & 12.61 & 13.03 & 15.14 & 12.35 & 8.13 & 15.69 & 12.34 & 16.32 & 14.57 & 13.07  \\
           ~$\bullet~$ Freecontrol (3 scenarios aug.) & \cmark & 0.77 & 1.81 & 1.98 & 2.19 & 3.08 & 3.96 & 3.29 & 0.52 & 2.55 & 3.76 & 3.10 & 4.10 & 5.37 & 2.81  \\
           ~$\bullet~$ Freecontrol (6 scenarios aug.) & \cmark & 6.81 & 7.54 & 7.67 & 5.54 & 5.45 & 5.10 & 2.19 & 1.31 & 3.63 & 8.59 & 5.35 & 8.86 & 6.26 & 5.72  \\
           \midrule
           ~$\bullet~$ DriveGEN (Only Snow aug.) & \cmark &  6.22 & 6.89 & 9.27 & 15.03 & 17.24 & 18.81 & 16.40 & 13.38 & 9.81 & 23.28 & 11.50 & 12.04 & 13.92 & 13.37   \\
           ~$\bullet~$ DriveGEN (3 scenarios aug.) & \cmark & 6.53 & 9.26 & 11.17 & 16.15 & \textbf{19.60} & \textbf{21.02} & \textbf{18.67} & 16.17 & 13.72 & \textbf{24.80} & 14.52 & 19.98 & \textbf{19.36} & 16.23  \\
           ~$\bullet~$ DriveGEN (6 scenarios aug.) & \cmark & 9.70 & \textbf{13.68} & \textbf{13.52} & \textbf{17.00} & 17.74 & 20.39 & 17.23 & \textbf{18.79} & \textbf{15.50} & 23.09 & \textbf{15.80} & \textbf{20.99} & 18.34 & \textbf{17.06}  \\     
         \bottomrule
         \end{tabular}
         }
    \end{center}
    \vspace{-0.05in}
\end{table*}

% Finally, we get the modified score estimate $\hat{\epsilon}_{\theta}$ through self-attention feature alignment $g_{sa}$ and shallow feature alignment $g_{sl}$ with classifier-free guidance~\cite{ho2022classifier}. XXXXXXXXXX

\section{Experiments}

%-------------------------------------------------------------------------------------------
% \vspace{-0.05in}
\begin{figure*}[t] 
  \centering
  \includegraphics[width=0.94\linewidth]{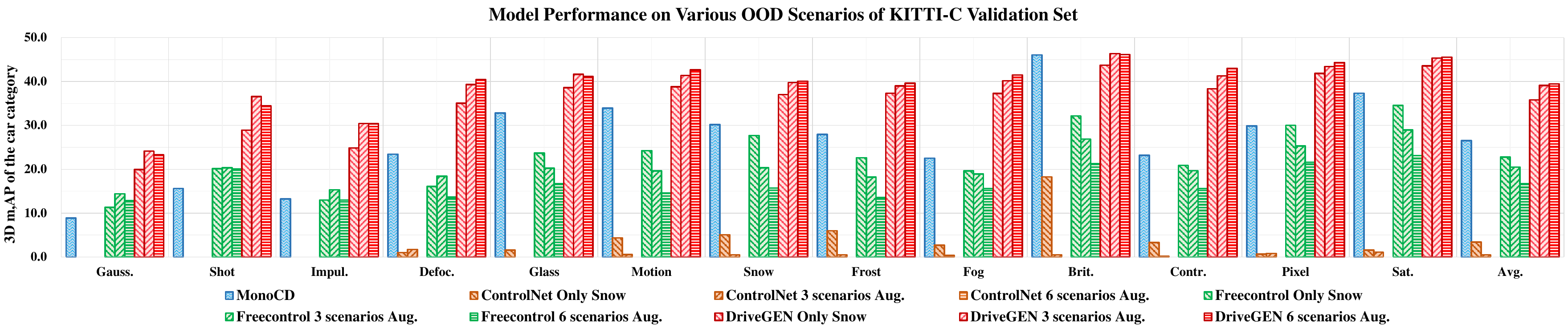} 
   \caption{Based on MonoCD~\cite{yan2024monocd}, we provide more comparisons with baselines on  KITTI-C, regarding {Mean $AP_{3D|R_{40}}$}.
   }
   \label{fig:exp_comp_fig}
   % \vspace{-0.07in}
\end{figure*}
%-------------------------------------------------------------------------------------------

\begin{table*}[t]
\setlength\tabcolsep{4pt}
    \begin{center}
    % \vspace{-0.12in}
    \caption{\label{tab:nus}
   Detection results on nuScenes-C and real-world scenarios of nuScenes, regarding mAP and NDS. 
   % More results of other scenarios are provided in Appendix \textbf{\emph{E}}.
    }
    \vspace{-0.05in}
    \scalebox{0.55}{
         \begin{tabular}{l|l|ccccccccc|ccc}
         \toprule
        \multirow{2}{*}{Metric} &  \multirow{2}{*}{Method} & \multicolumn{9}{c}{nuScenes-C} & \multicolumn{3}{c}{Real-world Scenarios} \\
         \cmidrule(lr){3-11} \cmidrule(lr){12-14} 
        & & Brightness & CameraCrash & ColorQuant & Fog & FrameLost & LowLight & MotionBlur & Snow  & Avg. & nuScenes-Night & nuScenes-Rainy & Avg.\\
         \midrule
        \multirow{2}{*}{mAP} & BEVFormer-tiny &  24.26  & 22.01  & 23.91  & 21.98  & 20.43  & 16.11  & 20.78  & 10.83  & 20.04 & 9.31  & 26.76 & 18.04  \\
        & ~$\bullet~$ DriveGEN (3k Snow) & \textbf{26.04}  & \textbf{24.13}  &\textbf{ 25.78 } & \textbf{24.13}  & \textbf{21.13}  & \textbf{17.20}  & \textbf{22.39}  & \textbf{11.72}  & \textbf{21.57}(\textcolor{red}{$\uparrow$1.53}) & \textbf{12.19}  & \textbf{29.11} & \textbf{20.65} (\textcolor{red}{$\uparrow$2.61})  \\ 
        \midrule
        \multirow{2}{*}{NDS} & BEVFormer-tiny &  34.93  & 33.06  & 34.56  & 33.04  & 31.42  & 28.42  & 31.81  & 23.27  & 31.31 &  17.92  & 37.35  & 27.64 \\
        & ~$\bullet~$ DriveGEN (3k Snow) &  \textbf{37.14}  & \textbf{35.34}  & \textbf{36.93}  & \textbf{35.35}  & \textbf{33.32}  & \textbf{30.42}  & \textbf{34.23}  & \textbf{24.09}  & \textbf{33.35}(\textcolor{red}{$\uparrow$2.04}) & \textbf{19.37}  & \textbf{40.95} & \textbf{30.16} (\textcolor{red}{$\uparrow$2.52}) \\ 
         \bottomrule
         \end{tabular}
         }
    \end{center}
    \vspace{-0.1in}
\end{table*}

We conduct experiments to validate monocular and multi-view 3D object detection, including four different training settings with various augmented OOD scenarios:
1) Augmented with traditional techniques (Traditional aug.);
2) Only augmented with Snow (Only Snow aug.);
3) Augmented with Snow, Rain and Fog (3 scenarios aug.);
4) Augmented with Snow, Rain, Fog, Night, Defocus and Sandstorm (6 scenarios aug.).
% More results are provided in Appendix \textbf{\emph{D}}.
% on KITTI~\cite{geiger2012we} and nuScenes~\cite{caesar2020nuscenes}

\noindent
\textbf{Datasets.}
For monocular 3d object detection, we split the images of KITTI~\cite{geiger2012we} into a training set (3712 images) and a validation set (3769 images) following the protocol from Monoflex~\cite{zhang2021objects}, encompassing three classes: Car, Pedestrian, and Cyclist.
We adopt the KITTI-C dataset~\cite{lin2025monotta} for validation, which includes 13 types of data corruptions on the validation set across four categories: Noise, Blur, Weather, and Digital~\cite{hendrycks2018benchmarking}.

As for multi-view 3D object detection, we first split real-world Night and Rainy validation scenarios based on descriptions of nuScenes~\cite{caesar2020nuscenes} following~\cite{liu2023bevfusion}. Next, we augment 500 daytime training scenes under the Snow condition. The augmented detectors are then evaluated on real-world Night and Rainy scenarios, as well as on the widely used Robo3D benchmark~\cite{xie2025benchmarking}. More details of datasets are provided in Appendix~\ref{sec:supp_dataset}.

\noindent\textbf{Implementation Details.}
We implement our method and other baselines in PyTorch~\cite{paszke2019pytorch}. Following Freecontrol~\cite{mo2024freecontrol}, we adopt keys from the self-attention of the first U-Net decoder layer as the features. In the self-prototype extraction stage,
we run DDIM inversion with 200 steps. Subsequently, we also run 200 steps of DDIM sampling in the prototype-guided diffusion stage. 
We set $s$ and $\tau$ to $7.5$ and $0.3$ following FreeControl~\cite{mo2024freecontrol}, while $\sigma$ is set to 0.1.
% All 3D detectors are trained by the mix of original and augmented data.
All 3D detectors are trained on a mix of original and augmented data.
All results in the manuscript are based on Stable Diffusion (SD) 1.5~\cite{rombach2022high}. More details and the results of SD 2.1 and XL 1.0 are put in Appendix~\ref{sec:supp_details}.

\noindent\textbf{Compared Methods.}
Based on typical or state-of-the-art (SOTA) 3D detectors~\cite{zhang2021objects,qin2022monoground,li2022bevformer,yan2024monocd},
we fully compare \ournet~with following methods: 
1) source-only, \ie directly apply the well-trained model to corrupted test data;
2) Traditional data augmentation, \ie Color Jitter and Brightness;
3) training-based controllable T2I diffusion: ControlNet~\cite{zhang2023adding} trains auxiliary modules for spatial control;
4) training-free controllable T2I diffusion: Freecontrol~\cite{mo2024freecontrol} enables zero-shot control of pretrained diffusion models. 

\noindent\textbf{Evaluation Protocols}.
We report the experimental results in the Average Precision (AP) for 3D bounding boxes, denoted as $AP_{3D|R_{40}}$.
On the KITTI-C dataset, the results present the mean values of three difficulty levels and Intersection over Union (IoU) thresholds are set to 0.7, 0.5, 0.5 for Cars and 0.5, 0.25, 0.25 for Pedestrians and Cyclists. 

As for nuScenes, the mean average precision (mAP) and nuScenes detection score (NDS) are calculated following BEVFormer~\cite{li2022bevformer}. The metrics also include five true positive metrics, including ATE, ASE, AOE, AVE, and AAE for measuring errors in translation, scale, orientation, velocity, and attributes.

% %-------------------------------------------------------------------------------------------
% % \vspace{-0.05in}
% \begin{figure*}[t] 
%   \centering
%   \includegraphics[width=\linewidth]{figures/fig_5.pdf} 
%   \vspace{-0.07in}
%    \caption{Qualitative comparisons on controllable T2I diffusion. The proposed DriveGEN achieves superior object preservation and OOD scenario generation in comparison to both training-based and training-free methods with Stable Diffusion 1.5~\cite{rombach2022high}. 
%    }
%    \label{fig:our_comparevis}
   
% \end{figure*}
% %-------------------------------------------------------------------------------------------

\subsection{Comparisons with Previous Methods}
% We first compare our \ournet~with previous methods on KITTI-C. 
% The results reported in Table~\ref{tab:kitti-c} and shown in Figure~\ref{fig:our_comparevis} give the following observations:
% 1) Directly applying the well-trained model to the corrupted test data (\ie source-only) suffers significant performance degradation due to the data distribution shifts.
% 2) Traditional data augmentation offers limited gains and often degrades model performance.
% 3) ControlNet fails to enhance 3d detectors, showing that fine-grained spatial control is challenging with limited training data. Note that we provide the segmentation masks~\cite{ravi2024sam} and rich prompts~\cite{chen2024internvl} for fine-tuning ControlNet (pre-trained on ade20k~\cite{zhou2017scene}).
% 4) Freecontrol brings a few gains in a few cases while leading to performance degradation in most scenarios, indicating that training-free T2I diffusion is a feasible solution but requires careful object preservation. 
% 5) \ournet~consistently outperforms all compared methods over all categories within various base models for both known and unseen scenarios, like improving MonoFlex by an average of {7.6 mAP} across 13 various OOD scenarios with the single Snow scenario.
% For nuScenes, even with only 500 augmented training scenes in the Night or Snow scenarios, the multi-view 3d detector~\cite{li2022bevformer} achieves a comprehensive improvement over NDS and mAP in the real-world task as shown in Table~\ref{tab:nus}.
For monocular 3d object detection, Table~\ref{tab:kitti-c} gives the following observations:
1) Applying the well-trained detector to corrupted scenarios suffers significant performance degradation due to the data distribution shifts and traditional data augmentation offers limited gains.
2) ControlNet (pre-trained on ade20k~\cite{zhou2017scene}) fails to enhance 3d detectors with limited training data even if it is provided with accurate segmentation masks~\cite{ravi2024sam} and rich prompts~\cite{chen2024internvl} for fine-tuning.
3) Freecontrol achieves minor gains in a few cases but generally degrades, indicating that training-free T2I diffusion is a feasible solution but requires careful object preservation.
4) \ournet~consistently outperforms all compared methods over all categories within various base models for both known and unseen scenarios, like improving MonoFlex by an average of {7.6 mAP} across 13 various OOD scenarios with the single Snow scenario augmentation.

As for multi-view 3D object detection, we randomly select 3k daytime training images and apply Snow augmentation (3k Snow) to enhance the detector~\cite{li2022bevformer}. Table~\ref{tab:nus} shows \ournet~consistently enhances BEVFormer-tiny across 8 OOD scenarios with an average of 1.53 mAP and 2.04 NDS, further confirming our effectiveness. 
Moreover, we validate the detector on real Night and Rainy scenarios of nuScenes. With the help of \ournet, it also achieves notable gains in real-world tasks with an average improvement of 2.61 mAP and 2.52 NDS, further demonstrating our superiority. More detailed results are put in the Appendix~\ref{sec:more_res}.

\begin{figure*}[t] 
  \centering
  \includegraphics[width=0.97\linewidth]{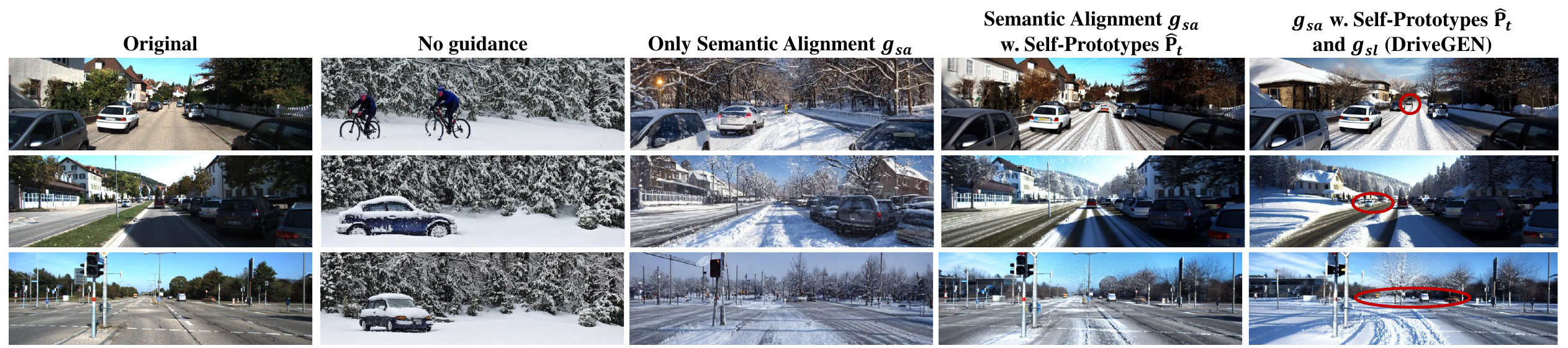} 
  \vspace{-0.05in}
   \caption{Ablation studies on semantic feature alignment loss $g_{sa}$, self-prototypes $\hat{\mathbf{P}}_t$ and
    shallow feature alignment loss $g_{sl}$. 
   }
   \label{fig:our_abla}
   
\end{figure*}
%-------------------------------------------------------------------------------------------

%-------------------------------------------------------------------------------------------
\begin{figure*}[t] 
  \centering
  \includegraphics[width=0.97\linewidth]{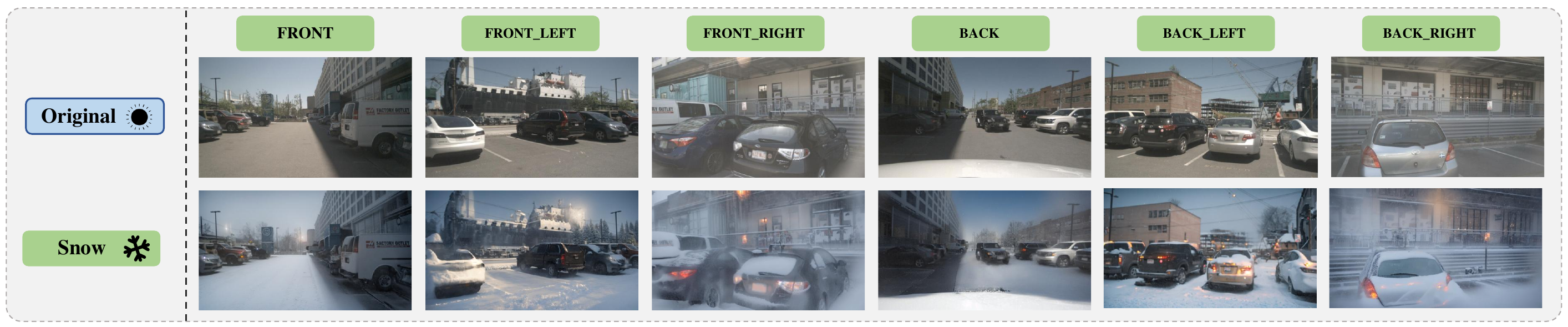} 
   \caption{Qualitative results of DriveGEN for multi-view 3D object detection on nuScenes. 
   \vspace{-0.05in}
   % DriveGEN supports vision-centric 3D detection tasks since our method only requires input images and corresponding object annotations without any additional diffusion model training.
   }
   \label{fig:our_vis}
   
\end{figure*}
%-------------------------------------------------------------------------------------------

\subsection{More Augmented Scenarios for Training}
An intuitive concern is whether the robustness of detectors progressively improves with the increasing of additional augmented scenarios (\ie Only Snow aug.$\rightarrow$3 scenarios aug.$\rightarrow$6 scenarios aug.). 
% To validate it, we provide more results based on the SOTA base monocular 3D detector~\cite{yan2024monocd} as shown in Figure~\ref{fig:exp_comp_fig}.
Based on Table~\ref{tab:kitti-c} and Figure~\ref{fig:exp_comp_fig}, we are given additional observations:
1) For ControlNet, the model performance drop for OOD scenarios grows as the number of augmentation scenarios increases, demonstrating that severe object omission leads to noisy training data.
2) For Freecontrol, the model performance across various OOD scenarios remains stable or tends to decline, indicating its object preservation may be unstable.
3) Owing to the precise object preservation, \ournet~obtains continuous overall performance (\ie Avg.) improvement across all categories with the incorporation of OOD scenarios. 
% which proves that the generalizability and robustness of these 3D detectors have been improved by \ournet.

\subsection{Ablation Studies}
To examine \ournet, we present qualitative results guided by different loss settings.
As shown in Figure~\ref{fig:our_abla}, compared with only generated via text prompts, introducing semantic-aware feature alignment $g_{sa}$ helps preserve more image structures.
Next, by guiding the diffusion process with our self-prototypes $\hat{\mathbf{P}}_t$, we achieve more accurate object geometry, alleviating potential object geometry loss.
Eventually, introducing our shallow feature alignment $g_{sl}$ further improves object preservation with precise 3D geometry, especially for small objects (highlighted by red circles). More ablation studies are provided in Appendix~\ref{sec:more_res}.

\subsection{Quantitative Results}
We provide visualizations for multi-view 3D object detection on nuScenes in Snow scenarios as shown in Figure~\ref{fig:our_vis}.
It is evident that \ournet~supports existing vision-centric 3D detection tasks since it requires only input images and corresponding object annotations without any additional diffusion model training, demonstrating that \ournet~achieves superior results even in challenging multi-view tasks. More visualizations are put in Appendix~\ref{sec:supp_vis}.

% our \ournet~supports existing vision-centric 3D detection tasks even though the input images are multi-views and contain multiple objects from different categories. 
% Moreover, our
% MonoTTA, it can produce 

\section{Conclusion}
\label{sec:conclu}
In this paper, we propose a method for robust 3D detection in driving via training-free controllable text-to-image diffusion generation, namely \ournet. Specifically, \ournet~ consists of: 1) Self-Prototype Extraction: To improve the guidance of self-attention features, we extract the self-prototypes by layouts to capture accurate object geometry, leveraging more precise features to guide diffusion.
2) Prototype-Guided Diffusion: To further preserve 3D object geometry, we conduct semantic-aware feature and shallow feature alignment during diffusion, alleviating the object misalignment and omission issues compared with previous methods.
Experiments on KITTI-C and nuScenes demonstrate the effectiveness of \ournet~in improving model robustness for vision-centric 3D detection.

% \noindent
% \textbf{Future directions.} 1) Our work focuses on images while future studies could explore videos for robust 3D detection in driving. 2) We explore  
\section*{Acknowledgements}
This work was supported by NSFC with Grant No. 62293482, by the Basic Research Project No. HZQB-KCZYZ-2021067 of Hetao Shenzhen HK S$\&$T Cooperation Zone, by Shenzhen General Program No. JCYJ20220530143600001, by Shenzhen-Hong Kong Joint Funding No. SGDX20211123112401002, by the Shenzhen Outstanding Talents Training Fund 202002, by Guangdong Research Project No. 2017ZT07X152 and No. 2019CX01X104, by the Guangdong Provincial Key Laboratory of Future Networks of Intelligence (Grant No. 2022B1212010001), by the Guangdong Provincial Key Laboratory of Big Data Computing, CHUK-Shenzhen, by the NSFC 61931024$\&$12326610, by the Key Area R$\&$D Program of Guangdong Province with grant No. 2018B030338001, by the Shenzhen Key Laboratory of Big Data and Artificial Intelligence (Grant No. ZDSYS201707251409055), by Shaanxi Mathematical Basic Science Research Project(No.23JSY047), and by Tencent $\&$ Huawei Open Fund.

% \clearpage
{
    \small
    \bibliographystyle{ieeenat_fullname}
    \bibliography{main}
}

\clearpage

\clearpage
\appendix
\setcounter{page}{1}
\maketitlesupplementary

\renewcommand{\thesection}{\Alph{section}}

In the supplementary, we first provide more related work and discussions to clarify existing vision-centric 3D object detection methods.
In addition, we provide more experimental details, visualizations, and results of \ournet. We organize our supplementary materials as follows.

\begin{itemize}
    \item In Appendix~\ref{sec:supp_related}, we review vision-centric 3D object detection and provide more discussions.
    \item In Appendix~\ref{sec:supp_dataset}, we provide more details of KITTI-C and the real-world scenarios of nuScenes.
    \item In Appendix~\ref{sec:supp_details}, we provide more details and results of \ournet~based on Stable Diffusion 2.1 and XL.
    \item In Appendix~\ref{sec:more_res}, we show more experimental results to demonstrate the effectiveness of the proposed \ournet.
    \item In Appendix~\ref{sec:supp_vis}, we show more qualitative results of our \ournet.
\end{itemize}

\section{More Related Work and Discussions}
\label{sec:supp_related}

In this section, we first provide more related work and discussions to clarify existing solutions to 3D object detection.

\noindent
\textbf{Vision-Centric 3D Object Detection.}
In autonomous driving, vision-centric 3D object detection is essential for accurate environment understanding. Traditional methods have often relied on LiDAR data, which provides precise depth information but necessitates costly hardware. Recently, there has been a shift toward using monocular and stereo cameras to reduce hardware dependency, but these methods struggle with depth accuracy, particularly at longer distances. Based on advances in transformer architectures, current approaches~\cite{ho2020denoising,dhariwal2021diffusion,song2021scorebased,rombach2022high,sohl2015deep,vincent2008extracting} attempt to bridge the gap between 2D images and 3D spatial reasoning using feature extraction and spatial alignment.
However, these models often demand high computational resources, making them challenging for real-time application. 
Diffusion-based models contribute to this task by offering a robust generation of 3D scene layouts capable of simulating diverse environments~\cite{yan2021videogpt,tulyakov2018mocogan}, while advances in multimodal integration~\cite{moon2022fusion,liu2022multimodal,li2022sensorfusion} enable the use of additional sensory data to enrich 3D object detection frameworks. The increasing trend of vision-centric 3D perception systems of autonomous vehicles further proves the effectiveness of model robustness.
% More and more 
% Additionally, diffusion-based models generate 3D scene layouts simulating diverse environments~\cite{yan2021videogpt,tulyakov2018mocogan}, enhancing scene diversity.

However, they still fall short in effectively capturing and aligning spatiotemporal features, which reduces overall accuracy and contextual understanding in complex environments. As discussed in the manuscript, existing 3D detectors often fail to maintain stable performance in OOD scenarios, which raises concerns about safety risks.

% yet frequently lacking fine-grained control over scene elements, which is critical in autonomous driving.
% necessary level of detail in critical scenarios. 
% Advances in multimodal integration~\cite{moon2022fusion,liu2022multimodal,li2022sensorfusion} incorporate data from multiple sources to enrich detection frameworks. 

%-------------------------------------------------------------------------------------------
\begin{figure}[t] 
  \centering
  \includegraphics[width=\linewidth]{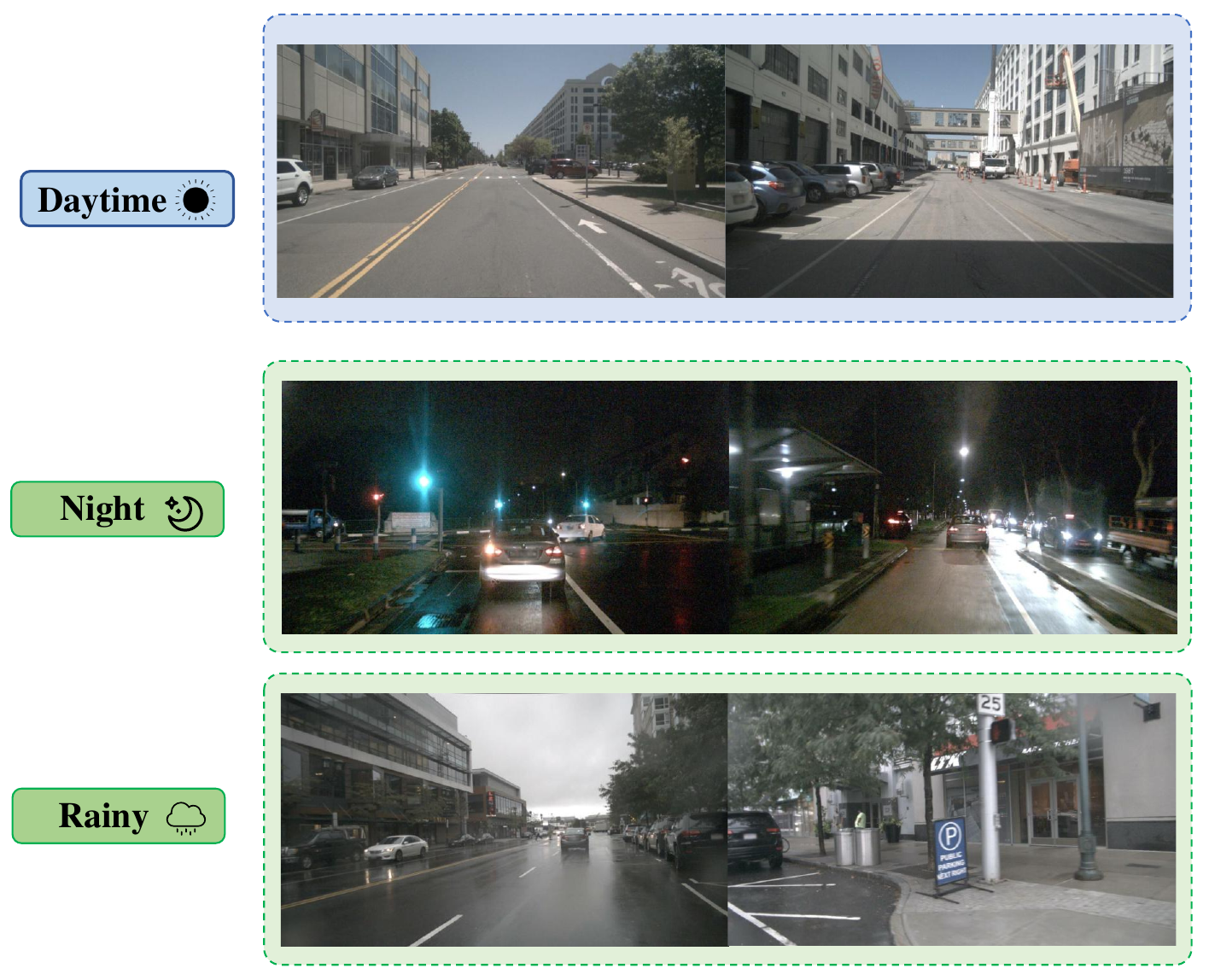} 
   \caption{Illustration of the real-world scenario Daytime, Night and Rainy of the nuScenes dataset. Given a pre-trained multi-view 3D detector, we enhance the detector with the augmented data from \ournet. Even if the augmented data never appears in the validation set (\eg Snow), \ournet~still improves the model performance, which shows the robustness and generalizability improvement of the augmented detector.
   }
   \label{fig:vis_nus_daytime2night}
\end{figure}
%-------------------------------------------------------------------------------------------

\section{More Details on Dataset Construction}
\label{sec:supp_dataset}

In this section, we first provide more visualizations of the KITTI-C dataset to illustrate the OOD scenarios.  Then, we offer more details of the real scenarios (\ie Daytime, Night and Rainy) of the nuScenes dataset.

\begin{figure*}[!h] 
  \centering
  \includegraphics[width=\linewidth]{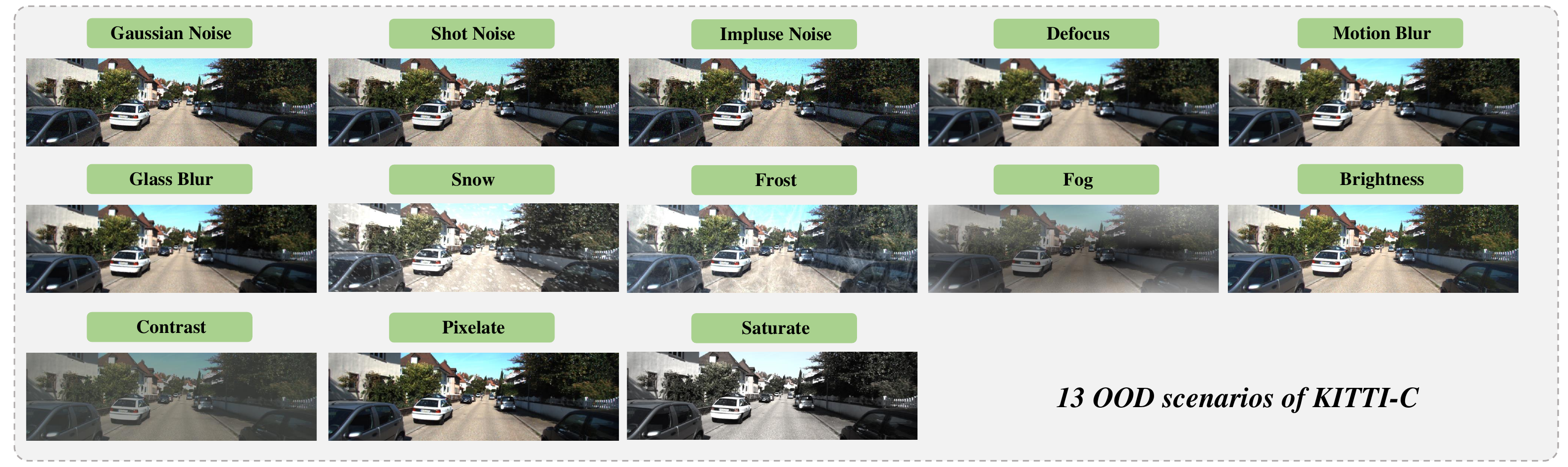} 
   \caption{Illustration of 13 distinct OOD scenarios of the KITTI-C~\cite{lin2025monotta} dataset.
   }
   \label{fig:vis_kittic}
\end{figure*}

For the KITTI-C dataset, as shown in Figure~\ref{fig:vis_kittic}, we follow MonoTTA to build 13 OOD scenarios based on the original KITTI validation set~\cite{lin2025monotta}, which is able to fully verify the effectiveness of each method in addressing dataset distribution shifts for Monocular 3D Object Detection.

As for the nuScenes dataset, we split images into Night and Rainy scenarios according to their descriptions, following~\cite{liu2023bevfusion}, as shown in Figure~\ref{fig:vis_nus_daytime2night}. 
To be specific, given a pre-trained multi-view 3D detector, \ournet~first augments the original training data into various OOD scenarios and then mixes the augmented data with the original training data for the model retraining.
It is worth mentioning that even if the augmented scenarios never appear in the validation set, \ournet~still consistently improves the model performance, demonstrating that \ournet~improves the robustness and generalizability of the augmented detector by injecting the knowledge from diffusion models.

%-------------------------------------------------------------------------------------------
%-------------------------------------------------------------------------------------------

%-------------------------------------------------------------------------------------------
\begin{figure*}[!h] 
  \centering
  \includegraphics[width=\linewidth]{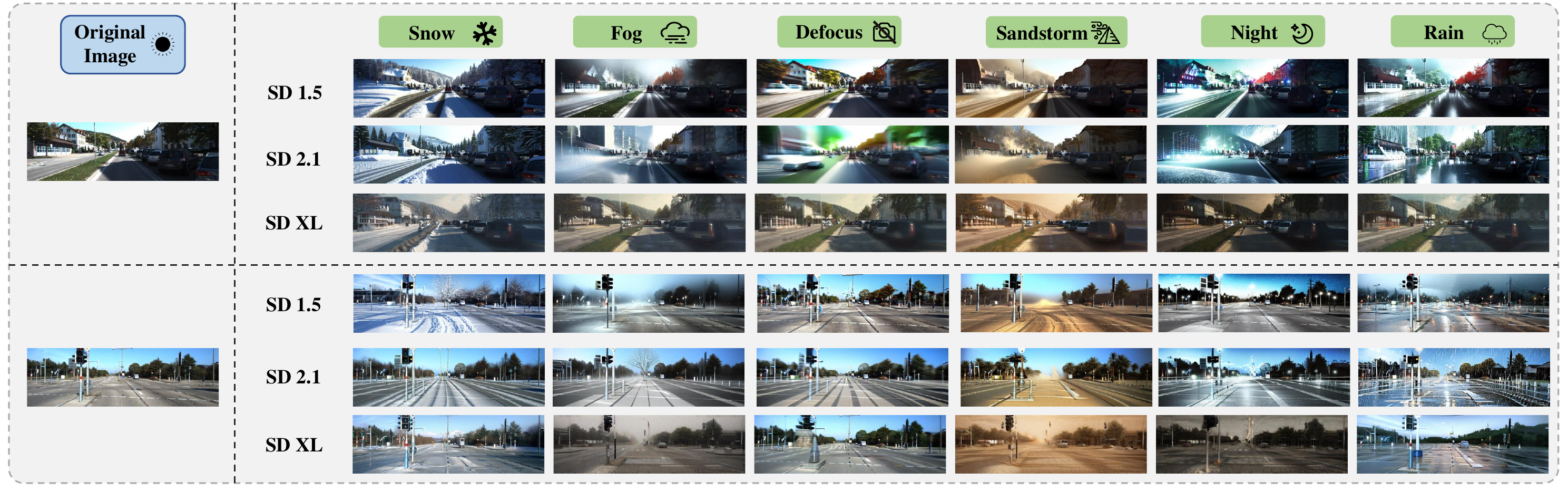} 
   \caption{Qualitative results of DriveGEN based on Stable Diffusion 2.1 and Stable Diffusion XL.
   }
   \label{fig:vis_sd21_xl}
\end{figure*}
%-------------------------------------------------------------------------------------------

%-------------------------------------------------------------------------------------------
\begin{figure*}[t] 
  \centering
  \includegraphics[width=\linewidth]{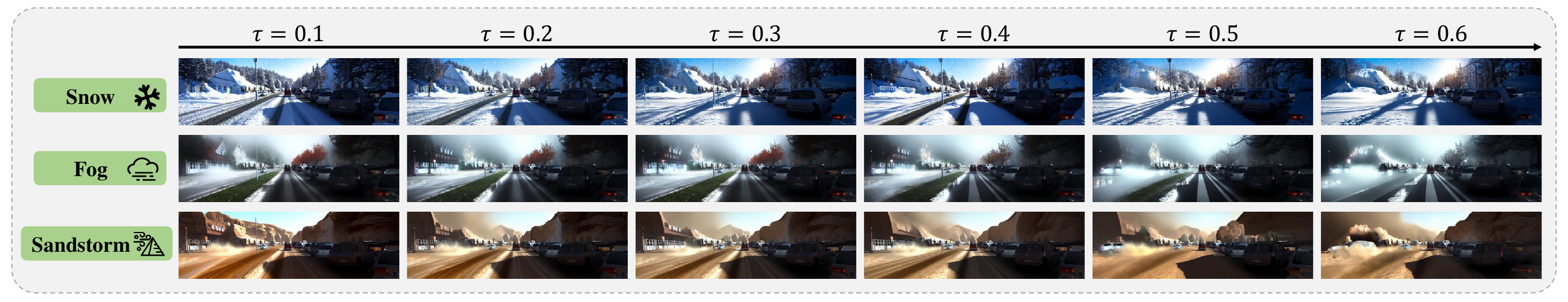} 
   \caption{\ournet~is able to control the severity of corruptions while still preserving all objects.
   }
   \label{fig:supp_abl_tau}
\end{figure*}
% \clearpage
%-------------------------------------------------------------------------------------------

\begin{figure}[!h]
% \vspace{-0.12in}
  \centering
  \includegraphics[width=\linewidth]{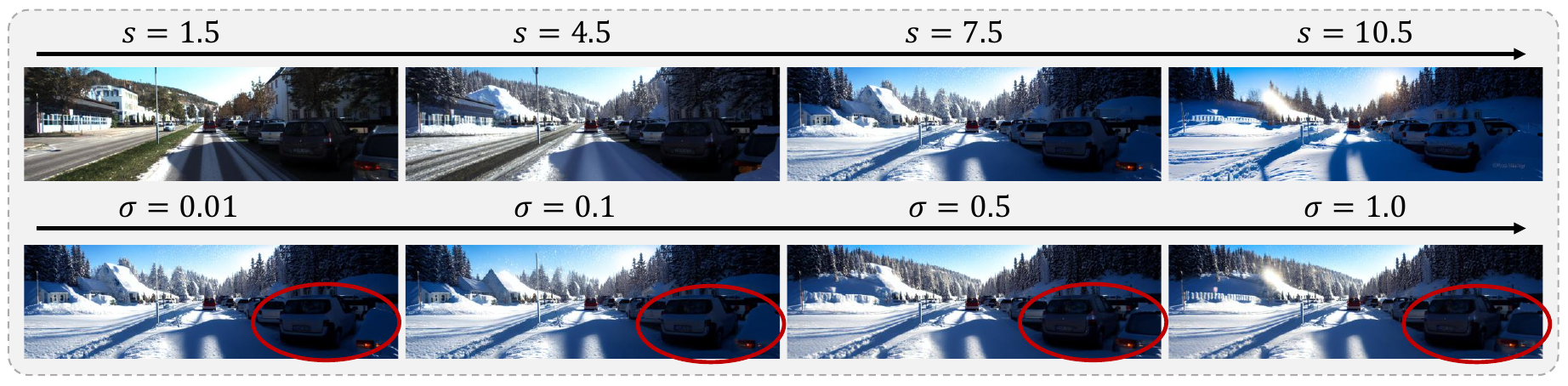} 
  \caption{More ablation studies of DriveGEN regarding hyper-parameters $s$ and $\sigma$.}
  \label{fig:abl_kitti}
  % \vspace{-0.15in}
\end{figure}

\section{More Details and Results with Stable Diffusion 2.1 and Stable Diffusion XL}
\label{sec:supp_details}
\noindent
\textbf{More Implementation Details.}
Based on PyTorch~\cite{paszke2019pytorch}, we conduct experiments with NVIDIA A100 (80GB of memory) GPUs and each method is executed on a single GPU.
To be specific, we adopt keys from the first self-attention layer of the U-Net decoder as the features following Freecontrol~\cite{mo2024freecontrol}. Besides, the number of steps for DDIM inversion and DDIM sampling is 200. We set $s$ and $\tau$ to $7.5$ and $0.3$ following existing method~\cite{mo2024freecontrol}, while $\sigma$ is set to 0.1. The corresponding ablation studies are put in appendix \textbf{D}.
All images of KITTI and nuScenes have the fixed generation size of $368 \times 1240$ and $896 \times 1600$, respectively. 

As for the training of 3D detectors, we follow their original settings without any hyper-parameter modification.
Specifically, given the augmented training data, we mix the original training data with all augmented OOD scenario for model training. Then, we validate all models on the original validation set to choose the best model because the original training data predominantly represents the most commonly occurring scenarios.  

\noindent
\textbf{Stable Diffusion 2.1 and XL 1.0}
In this section, we present additional quantitative results based on Stable Diffusion 2.1 and Stable Diffusion XL. Since \ournet~requires no additional diffusion model training, it is straightforward to extend \ournet~to U-Net architecture~\cite{ronneberger2015u} based diffusion model, as shown in Figure~\ref{fig:vis_sd21_xl}. 
These results demonstrate that \ournet~effectively supports Stable Diffusion models across different versions (\ie U-Net-based), showcasing its flexibility and scalability for integrating with vision-centric 3D detection methods.
% existing vision-centric 3D detection tasks since it requires only input images and corresponding object annotations without any additional diffusion model training, demonstrating that \ournet~can achieve superior results even in challenging multi-view tasks.

\begin{table*}[!h]
\setlength\tabcolsep{7pt}
\renewcommand\arraystretch{1.0}
    \begin{center}
    \caption{\label{tab:abl_supp1} Comparison on the {KITTI-C} dataset, severity \textbf{level 1} regarding {Mean $AP_{3D|R_{40}}$}. Each scenario represents the training of the 3D detector, which is enhanced with corresponding OOD data.
    The \textbf{bold} number indicates the best result. 
    % \textcolor{blue}{Blue} cells show scenarios with similar training augmentation data for the baseline while other cells represent unseen scenarios.
    }  
    % \vspace{-0.08in}
    \scalebox{0.65}{
         \begin{tabular}{l|ccc|ccc|cccc|ccc|c}
         \toprule
         \multicolumn{15}{c}{\textbf{\textbf{Car}, IoU @ 0.7, 0.5, 0.5}} \\
         \midrule
         \multirow{2}{*}{Method}  &
         \multicolumn{3}{c|}{Noise} & 
         \multicolumn{3}{c|}{Blur} & 
         \multicolumn{4}{c|}{Weather} & 
         \multicolumn{3}{c|}{Digital} & 
         \multirow{2}{*}{Avg.} \\
        \cmidrule(lr){2-4} \cmidrule(lr){5-7} \cmidrule(lr){8-11} \cmidrule(lr){12-14} 
        & Gauss. & Shot & Impul. & Defoc. & Glass & Motion & Snow & Frost & Fog & Brit. & Contr. & Pixel & Sat. \\
         \midrule
        Monoflex~\cite{zhang2021objects} &  13.06 &  20.91 &  14.09 &  20.17 &  28.59 &  30.34 &  33.64 &  30.31 &  19.58 &  45.22 &  20.01 &  29.07 &  38.85 & 26.45 \\
        \midrule
    
        ~$\bullet~$ Snow  & 17.07 & 26.78 & 23.78 & 32.89 & 37.52 & 39.06 & {40.61} & 34.91 & 25.29 & {46.21} & 27.12 & 38.25 & 44.45 & 33.38 \\  
        ~$\bullet~$ Fog  & 17.98 & 29.72 & 20.66 & 34.10 & 39.25 & 38.47 & 39.49 & 38.48 & 30.05 & 47.90 & 30.71 & 37.02 & 45.19 & 34.54  \\
        ~$\bullet~$ Rainy  & 19.49 & 31.10 & \textbf{27.44} & 33.34 & \textbf{39.55} & 39.41 & \textbf{41.42} & \textbf{41.10} & \textbf{37.11} & 47.59 & \textbf{38.69} & 41.05 & 45.35 & 37.13 \\
        ~$\bullet~$ Night   & 23.28 & \textbf{35.21} & 26.82 & \textbf{35.13} & 39.24 & 39.62 & 40.69 & 40.77 & 34.68 & 47.46 & 36.06 & \textbf{42.40} & \textbf{45.55} & \textbf{37.46} \\
        ~$\bullet~$ Defocus  & 22.06 & 32.16 & 26.64 & 34.45 & 37.71 & \textbf{40.69} & 40.37 & 37.94 & 31.63 & \textbf{48.49} & 34.88 & 41.19 & 44.23 & 36.34 \\
        ~$\bullet~$ Sandstorm   & \textbf{24.46} & 33.96 & 26.09 & 31.50 & 37.18 & 37.07 & 40.79 & 38.83 & 32.65 & 43.85 & 33.44 & 37.98 & 42.60 & 35.41 \\
         \bottomrule
         \end{tabular}
         }
    \end{center}
    % \vspace{-0.15in}
\end{table*}

\begin{table*}[!h]
\setlength\tabcolsep{7pt}
\renewcommand\arraystretch{1.0}
    \begin{center}
    \caption{\label{tab:abl_supp2} Comparison on the {KITTI-C} dataset, severity \textbf{level 1} regarding {Mean $AP_{3D|R_{40}}$}. Each setting represents the training of the 3D detector is enhanced with corresponding mixed OOD data.
    The \textbf{bold} number indicates the best result. 
    }  
    % \vspace{-0.08in}
    \scalebox{0.63}{
         \begin{tabular}{l|ccc|ccc|cccc|ccc|c}
         \toprule
         \multicolumn{15}{c}{\textbf{\textbf{Car}, IoU @ 0.7, 0.5, 0.5}} \\
         \midrule
         \multirow{2}{*}{Method}  &
         \multicolumn{3}{c|}{Noise} & 
         \multicolumn{3}{c|}{Blur} & 
         \multicolumn{4}{c|}{Weather} & 
         \multicolumn{3}{c|}{Digital} & 
         \multirow{2}{*}{Avg.} \\
        \cmidrule(lr){2-4} \cmidrule(lr){5-7} \cmidrule(lr){8-11} \cmidrule(lr){12-14} 
        & Gauss. & Shot & Impul. & Defoc. & Glass & Motion & Snow & Frost & Fog & Brit. & Contr. & Pixel & Sat. \\
         \midrule
        Monoflex~\cite{zhang2021objects} &  13.06 &  20.91 &  14.09 &  20.17 &  28.59 &  30.34 &  33.64 &  30.31 &  19.58 &  45.22 &  20.01 &  29.07 &  38.85 & 26.45 \\
        \midrule
         ~$\bullet~$ Snow \& Fog \& Rain & {25.63} & {37.04} & {29.13} & \textbf{{34.13}} & \textbf{{39.15}} & 36.81 & 38.58 & 37.98 & 33.93 & 45.39 & 34.66 & 39.36 & 43.81 & 36.58 \\
        ~$\bullet~$ Defocus \& Night \& Sandstorm & \textbf{28.93} & \textbf{37.91} & \textbf{32.12} & 31.59 & 37.66 & \textbf{39.27} & \textbf{40.34} & \textbf{40.88} & \textbf{37.72} & \textbf{47.15} & \textbf{38.00} & \textbf{42.34} & \textbf{45.58} & \textbf{38.42}  \\ 
         \bottomrule
         \end{tabular}
         }
    \end{center}
    \vspace{-0.08in}
\end{table*}

% \begin{table}[t]
% \setlength\tabcolsep{4pt}
% % \renewcommand\arraystretch{0.85}
%     \begin{center}
%     \caption{\label{tab:nus_detail}
%    Detailed results of Daytime $\rightarrow$ Night on nuScenes.
%     }
%     \scalebox{0.6}{
%          \begin{tabular}{l|cccccccc}
%          \toprule
%         Method & NDS$\uparrow$ & mAP$\uparrow$  & mATE$\downarrow$ & mASE$\downarrow$ & mAOE$\downarrow$ & mAVE$\downarrow$ & mAAE$\downarrow$
%          \\        
%          \midrule
%          BEVFormer-small~\cite{li2022bevformer} & 0.228 & 0.173  & 0.860 & 0.498 & 0.659 & 0.992 & 0.586 \\
%          % \midrule
%         % ~$\bullet~$ DriveGEN & \\
%         ~$\bullet~$ DriveGEN (Night aug.) & {0.256} & {0.187} & \textbf{0.800} & 0.487 & 0.653 & 0.838 & 0.605   \\
%         ~$\bullet~$ Oracle (Daytime \& Night) & \textbf{0.273} & \textbf{0.196}  & 0.826 & \textbf{0.479} & \textbf{0.575} & \textbf{0.771} & \textbf{0.602}  \\
%          \bottomrule
%          \end{tabular}
%          }
%     \end{center}
%     \vspace{-0.1in}
% \end{table}

\begin{table*}[t]
\setlength\tabcolsep{7pt}
\renewcommand\arraystretch{1.0}
    \begin{center}
    \caption{\label{tab:kitti-c-cyc} Comparison on the Cyclist category of the {KITTI-C} dataset regarding {Mean $AP_{3D|R_{40}}$}. \textbf{Bold} number indicates the best result. 
    }  
    % \vspace{-0.08in}
    \scalebox{0.63}{
         \begin{tabular}{lc|ccc|ccc|cccc|ccc|c}
         \toprule
         \multicolumn{15}{c}{\textbf{\textbf{Cyclist}, IoU @ 0.7, 0.5, 0.5}} \\
         \midrule
         \multirow{2}{*}{Method}  &
         \multirow{2}{*}{\shortstack{Training-free\\diffusion}} &
         \multicolumn{3}{c|}{Noise} & 
         \multicolumn{3}{c|}{Blur} & 
         \multicolumn{4}{c|}{Weather} & 
         \multicolumn{3}{c|}{Digital} & 
         \multirow{2}{*}{Avg.} \\
        \cmidrule(lr){3-5} \cmidrule(lr){6-8} \cmidrule(lr){9-12} \cmidrule(lr){13-15} 
        & & Gauss. & Shot & Impul. & Defoc. & Glass & Motion & Snow & Frost & Fog & Brit. & Contr. & Pixel & Sat. \\
         \midrule
        
        Monoflex~\cite{zhang2021objects} &  & 0.43 & 2.41 & 0.64 & 2.76 & 8.30 & 9.14 & 12.85 & 11.09 & 5.73 & 17.44 & 4.84 & 3.25 & 9.89 & 6.83  \\
        \midrule
        
        ~$\bullet~$ Color Jitter (Traditional aug.) & & 0.63 & 3.15 & 1.91 & 1.62 & 3.43 & 7.92 & 11.03 & 10.09 & 4.60 & 12.41 & 4.61 & 1.43 & 10.23 & 5.62  \\
        ~$\bullet~$ Brightness (Traditional aug.) & & 0.21 & 1.16 & 0.25 & 1.33 & 3.45 & 6.14 & 9.67 & 8.81 & 4.89 & 13.66 & 5.82 & 2.02 & 7.93 & 5.03  \\
        
        \midrule
        ~$\bullet~$ ControlNet (Only Snow aug.) & \xmark &  0.00 & 0.30 & 0.00 & 0.00 & 3.77 & 4.29 & 7.27 & 6.47 & 6.97 & 15.79 & 6.49 & 1.67 & 2.54 & 4.27 \\  
        ~$\bullet~$ ControlNet (3 scenarios aug.) & \xmark &  0.00 & 0.00 & 0.00 & 0.00 & 0.00 & 0.00 & 0.00 & 2.50 & 1.50 & 1.82 & 1.35 & 0.00 & 0.00 & 0.55  \\  
        ~$\bullet~$ ControlNet (6 scenarios aug.) & \xmark & 0.00 & 0.00 & 0.00 & 0.00 & 0.00 & 0.00 & 0.00 & 0.00 & 0.00 & 0.00 & 0.00 & 0.00 & 0.00 & 0.00 \\  
         \midrule
        ~$\bullet~$ Freecontrol (Only Snow aug.) & \cmark & 1.58 & 4.43 & 1.72 & 0.00 & 0.39 & 0.94 & 3.97 & 1.52 & 0.54 & 5.26 & 0.68 & 1.07 & 7.50 & 2.28   \\  
        ~$\bullet~$ Freecontrol (3 scenarios aug.) & \cmark & 0.00 & 0.00 & 0.00 & 0.00 & 2.50 & 0.00 & 1.25 & 1.04 & 1.91 & 3.82 & 1.78 & 0.00 & 2.47 & 1.14 \\  
        ~$\bullet~$ Freecontrol (6 scenarios aug.) & \cmark & 0.19 & 0.24 & 0.45 & 0.00 & 0.31 & 0.00 & 0.55 & 0.52 & 1.01 & 2.13 & 2.12 & 0.61 & 0.81 & 0.69 \\  
         \midrule
        ~$\bullet~$ DriveGEN (Only Snow aug.) & \cmark & 0.70 & 1.27 & 0.61 & 1.34 & 5.26 & 5.27 & 10.90 & 7.12 & 3.73 & 15.14 & 4.37 & 1.74 & 11.24 & 5.28  \\  
        ~$\bullet~$ DriveGEN (3 scenarios aug.) & \cmark & 1.04 & 3.42 & 1.53 & 2.36 & 5.62 & 7.69 & 8.14 & 4.17 & 5.20 & 13.24 & 5.23 & 4.58 & 11.07 & 5.64  \\  
        ~$\bullet~$ DriveGEN (6 scenarios aug.) & \cmark & 0.53 & 0.93 & 0.54 & 3.07 & 10.95 & 9.38 & 11.12 & 12.60 & 9.07 & 15.39 & 10.81 & 1.99 & 8.05 & \textbf{7.26}  \\ 
        
         \midrule
        \midrule

        MonoGround~\cite{qin2022monoground} &  & 0.21 & 1.86 & 1.34 & 0.83 & 2.93 & 2.23 & 5.00 & 3.43 & 0.94 & 11.48 & 1.21 & 2.04 & 5.92 & 3.03 \\
        \midrule
        
        ~$\bullet~$ Color Jitter (Traditional aug.) & & 0.39 & 2.67 & 2.11 & 0.31 & 2.03 & 2.19 & 5.38 & 4.63 & 1.12 & 13.64 & 1.67 & 2.89 & 5.00 & 3.39  \\
        ~$\bullet~$ Brightness (Traditional aug.) & & 0.06 & 0.61 & 0.22 & 0.36 & 1.33 & 1.06 & 4.72 & 2.32 & 1.41 & 6.87 & 0.78 & 0.90 & 2.81 & 1.80  \\
        
        \midrule
        ~$\bullet~$ ControlNet (Only Snow aug.) & \xmark &  0.00 & 0.00 & 0.52 & 0.00 & 0.77 & 1.33 & 0.44 & 1.10 & 0.14 & 6.77 & 0.30 & 0.50 & 0.54 & 0.95 \\  
        ~$\bullet~$ ControlNet (3 scenarios aug.) & \xmark &  0.00 & 0.00 & 0.00 & 0.00 & 0.30 & 0.00 & 0.83 & 1.50 & 0.00 & 1.70 & 0.00 & 0.37 & 0.00 & 0.36  \\  
        ~$\bullet~$ ControlNet (6 scenarios aug.) & \xmark & 0.00 & 0.00 & 0.00 & 0.00 & 0.00 & 0.00 & 0.00 & 0.00 & 0.00 & 0.00 & 0.00 & 0.00 & 0.00 & 0.00  \\  
         \midrule
        ~$\bullet~$ Freecontrol (Only Snow aug.) & \cmark & 0.46 & 0.70 & 0.32 & 1.07 & 0.17 & 0.50 & 1.60 & 0.91 & 0.21 & 4.48 & 0.17 & 1.93 & 4.50 & 1.31   \\  
        ~$\bullet~$ Freecontrol (3 scenarios aug.) & \cmark & 0.19 & 0.33 & 0.43 & 0.00 & 0.52 & 0.50 & 0.51 & 0.33 & 0.58 & 1.32 & 0.42 & 0.38 & 0.38 & 0.45  \\  
        ~$\bullet~$ Freecontrol (6 scenarios aug.) & \cmark & 0.00 & 0.34 & 0.62 & 0.00 & 0.42 & 0.56 & 0.00 & 0.00 & 1.00 & 2.07 & 0.83 & 1.25 & 0.74 & 0.60  \\  
         \midrule
        ~$\bullet~$ DriveGEN (Only Snow aug.) & \cmark & 0.13 & 0.81 & 0.38 & 0.31 & 2.23 & 3.66 & 3.96 & 2.02 & 0.90 & 8.46 & 1.74 & 2.07 & 4.58 & 2.40  \\  
        ~$\bullet~$ DriveGEN (3 scenarios aug.) & \cmark &  0.61 & 2.17 & 1.51 & 1.09 & 3.24 & 2.73 & 5.57 & 5.30 & 1.05 & 9.79 & 1.33 & 4.66 & 6.10 & 3.47  \\  
        ~$\bullet~$ DriveGEN (6 scenarios aug.) & \cmark & 1.49 & 2.16 & 1.66 & 3.30 & 5.97 & 5.55 & 5.64 & 5.49 & 2.49 & 9.37 & 3.48 & 3.79 & 5.65 & \textbf{4.31}  \\ 

         \bottomrule
         \end{tabular}
         }
    \end{center}
    \vspace{-0.05in}
\end{table*}

\begin{table*}[!h]
\setlength\tabcolsep{8pt}
    \begin{center}
    % \vspace{-0.12in}
    \caption{\label{tab:nus-small}
   Detection results on nuScenes-C and real-world scenarios of nuScenes, regarding mAP and NDS. 
   % More results of other scenarios are provided in Appendix \textbf{\emph{E}}.
    }
    \vspace{-0.05in}
    \scalebox{0.63}{
         \begin{tabular}{l|l|ccccccccc|ccc}
         \toprule
        \multirow{2}{*}{Metric} &  \multirow{2}{*}{Method} & \multicolumn{9}{c}{nuScenes-C} & {Real-world Scenarios} \\
         \cmidrule(lr){3-11} \cmidrule(lr){12-12} 
        & & Brightness & CameraCrash & ColorQuant & Fog & FrameLost & LowLight & MotionBlur & Snow  & Avg. & nuScenes-Night\\
         \midrule
        \multirow{2}{*}{mAP} & BEVFormer-small &  36.12 & 23.25 & 36.05 & 32.70 & \textbf{32.16} & 23.61 & \textbf{32.03} & 13.66 & 28.70 &19.59  \\
        & ~$\bullet~$ DriveGEN (3k Snow) &  \textbf{37.99} & \textbf{24.20} & \textbf{37.66} & \textbf{34.60} & {31.74} & \textbf{25.82} & \textbf{32.89} & \textbf{17.63} & \textbf{30.32} & \textbf{22.39}  \\ 
        \midrule
        \multirow{2}{*}{NDS} & BEVFormer-small &  47.36 & 39.49 & 47.28 & 45.01 & \textbf{44.05} & 38.62 & 44.46 & 28.45 & 41.84 & 27.27  \\
        & ~$\bullet~$ DriveGEN (3k Snow) & \textbf{48.90} & \textbf{40.44} & \textbf{48.63} & \textbf{46.56 }& 43.72 & \textbf{39.67} & \textbf{45.47} & \textbf{32.32} & \textbf{43.21} & \textbf{28.86}  \\
         \bottomrule
         \end{tabular}
         }
    \end{center}
    \vspace{-0.05in}
\end{table*}

\section{More Experimental Results}
\label{sec:more_res}
In this part, we conduct ablation studies to show the effectiveness of \ournet, like enhancing training data via various single OOD scenarios as shown in Table~\ref{tab:abl_supp1}, as well as enhancing training data by three additional OOD scenarios as shown in Table~\ref{tab:abl_supp2}. 
Meanwhile, we provide more detailed experimental results of the Cyclist category and the results regarding Moderate $AP_{3D|R_{40}}$ on the KITTI-C dataset.
Eventually, we provide more results of nuScenes with the BEVFormer-small~\cite{li2022bevformer}.

\noindent
\textbf{Ablation Study on Augmented OOD Scenarios.}
To fully validate the effectiveness of \ournet, we provide more results of enhancing training data via various single OOD scenarios as shown in Table~\ref{tab:abl_supp1} and enhancing training data by three additional OOD scenarios as shown in Table~\ref{tab:abl_supp2}. 
On the one hand, compared with the Snow augmentation. setting in the manuscript, \ournet~improves the 3D detection model with stable performance improvement by the other 5 OOD scenarios. In Table~\ref{tab:abl_supp1}, even with a single augmentation, \ournet~improves the 3D detector~\cite{zhang2021objects} up to \textbf{11.01 mAP} (\ie Night) across 13 OOD scenarios.
On the other hand, if we enhance the 3D detector with another three scenarios (\ie Defocus, Night, Sandstorm),
\ournet~still achieves significant performance improvement compared with the pre-trained 3D detector as shown in Table~\ref{tab:abl_supp2}.

\noindent
\textbf{Ablation Study on Hyper-parameters.}
One intuitive concern is whether the severity of corruptions can be controlled by \ournet~while still preserving all annotated objects with precise geometry. To this end, we provide more qualitative results regarding various values of $\tau$ as shown in Figure~\ref{fig:supp_abl_tau}. 
It is clearly observed that the corresponding corruption progressively exerts a more severe impact on the background with the increasing of $\tau$. However, \ournet~maintains all objects well with their precise 3D geometry, thereby demonstrating the effectiveness of the proposed method.
Meanwhile, we also analyze the effects of $s$ and $\sigma$ as shown in Figure~\ref{fig:abl_kitti}. 
As $s$ increases, corruptions intensify while larger $\sigma$ retains more original object details. 
It also shows that \ournet~consistently preserves objects across various settings, validating its insensitivity to hyperparameters.

To summarize, \ournet~enhances the generalizability and robustness of vision-centric 3D detectors with diverse augmented OOD scenarios, achieving stable performance improvement within various training data settings. These experimental results further demonstrate the effectiveness of \ournet.

\noindent
\textbf{Model Performance on Cyclist.}
We further provide more results of the cyclist category of the KITTI-C dataset as shown in Table~\ref{tab:kitti-c-cyc}. As mentioned in MonoTTA~\cite{lin2025monotta}, even Fully Test-Time Adaptation methods (\ie allowed to access test data for model adaptation ) only gain limited performance improvement on the Cyclist category.
Table~\ref{tab:kitti-c-cyc} gives a similar observation: even if \ournet~achieves the best performance in these extremely difficult cases, all methods only gain limited performance improvement, which indicates that 
the challenge of minority-class object detection still requires further investigation.

\noindent
\textbf{Model Performance Regarding the Moderate Level.}
In 3D object detection, the performance for the \emph{Moderate} difficulty level of the KITTI dataset is one of the most significant indicators of model effectiveness.
To this end, we provide more experimental results as shown in Table~\ref{tab:kitti-c-moderate}.
This table shows that \ournet~still achieves the best average performance within various augmentation settings and base models, demonstrating the effectiveness of our method.

\noindent
\textbf{More Results on NuScenes and NuScenes-C.}
Based on BEVFormer-small~\cite{li2022bevformer}, we apply the Snow augmentation (3k Snow) to enhance the detector as mentioned in the manuscript.
Table~\ref{tab:nus-small} shows that \ournet~consistently enhances BEVFormer-small across 8 OOD scenarios with an average of 1.62 mAP and 1.37 NDS, and across the real-world scenario of nuScenes (\ie Night) with 2.80 mAP and 1.59 NDS.
These results further demonstrate our effectiveness and superiority.

% \noindent
% \textbf{Detailed discussions on nuScenes.}

% \clearpage
%-------------------------------------------------------------------------------------------
\begin{figure*}[t] 
  \centering
  \includegraphics[width=\linewidth]{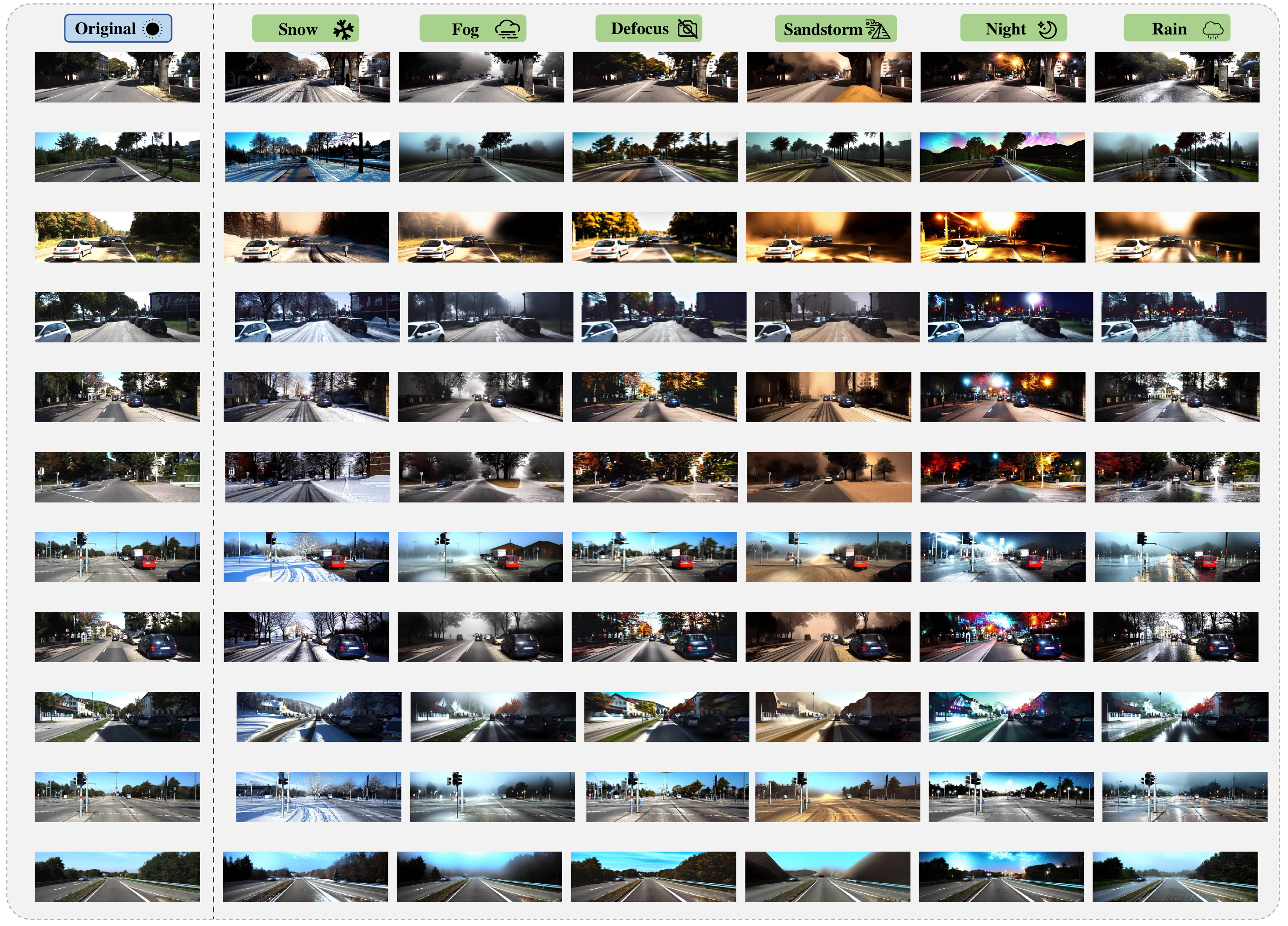} 
  \caption{More qualitative results of DriveGEN for the training data of the KITTI dataset.}
  \label{fig:vis_kitti}
\end{figure*}
% \clearpage
%-------------------------------------------------------------------------------------------

\section{More Qualitative Results}
\label{sec:supp_vis}
\begin{figure*}[t] 
  \centering
  \includegraphics[width=\linewidth]{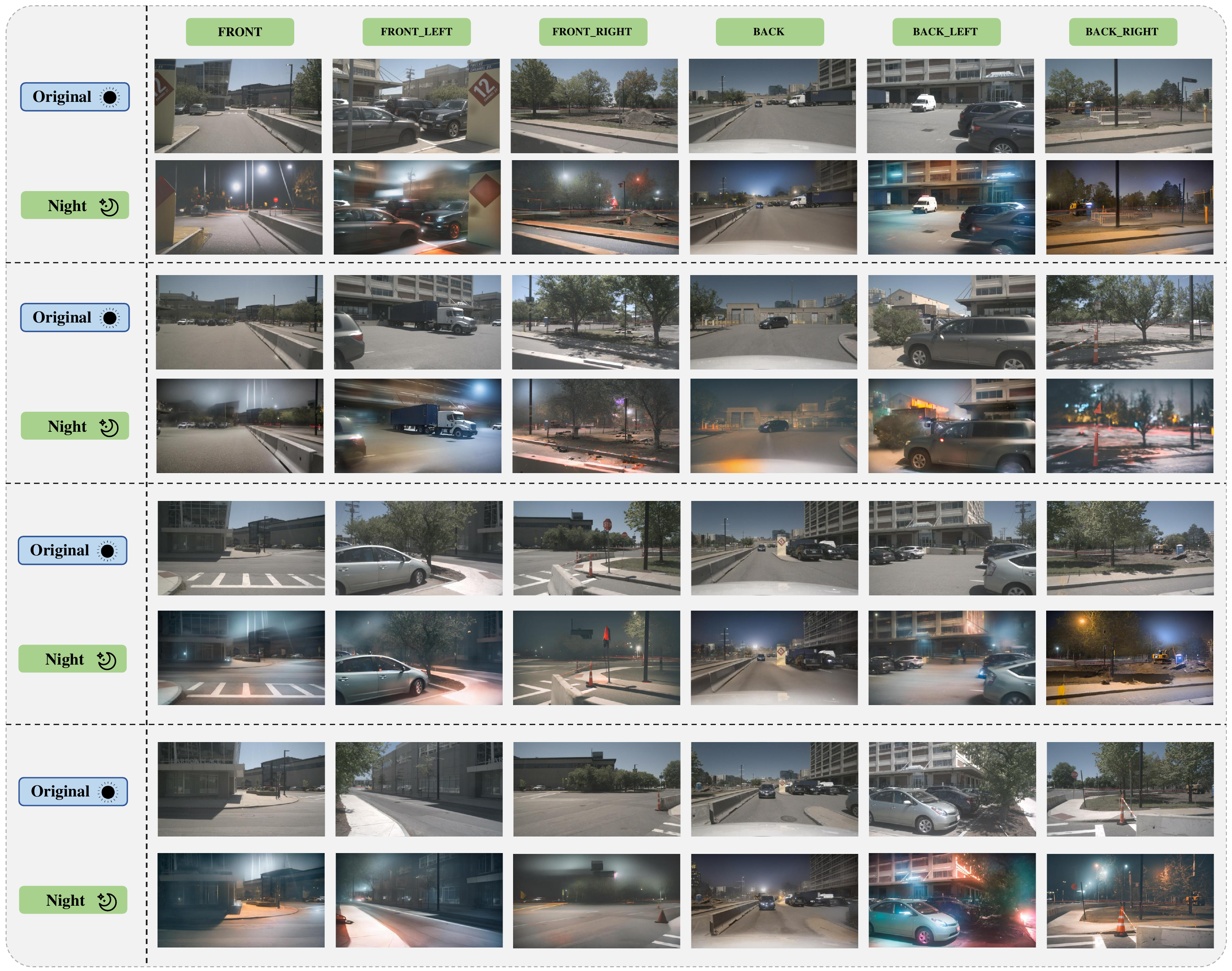} 
   \caption{More qualitative results of DriveGEN for multi-view training images of the nuScenes dataset.
   }
   \label{fig:vis_nus}
\end{figure*}

As shown in Figure~\ref{fig:vis_kitti}, we first provide more qualitative results based on the training images of the KITTI dataset. In addition, we also offer more qualitative results based on the training images of the nuScenes dataset as shown in Figure~\ref{fig:vis_nus}.
It is evident that \ournet~supports existing vision-centric 3D detection tasks (\ie monocular 3D object detection and multi-view 3D object detection) since our method only requires input images and corresponding object annotations without any additional diffusion model training, demonstrating that \ournet~can preserve all objects and enhance training data, thus achieving superior results even in challenging multi-view tasks.

\begin{table}[!h]
\setlength\tabcolsep{4pt}
    \begin{center}
    % \vspace{-0.12in}
    \caption{\label{tab:PSNR_SSIM} The quality comparisons of object regions based on PSNR and SSIM.}
   %  }
    \scalebox{0.6}{
         \begin{tabular}{c|l|ccccccccc}
         \toprule
        Metric& Method & Defocus & Snow & Fog & SandStorm & Night & Rainy & Avg. \\
         \midrule
        \multirow{3}{*}{PSNR} & ControlNet & 7.671  & 7.551  & 8.130  & 7.764  & 7.825  & 8.129  & 7.845  \\
        & FreeControl &  11.883  & 10.601  & 12.529  & 11.957  & 11.515  & 12.119  & 11.767 \\ 
        & DriveGEN &  \textbf{19.584}  & \textbf{18.963}  & \textbf{19.528}  & \textbf{19.551}  & \textbf{18.906}  & \textbf{19.308}  & \textbf{19.306} \\ 
        \midrule
        \multirow{3}{*}{SSIM} & ControlNet & 0.077  & 0.069  & 0.081  & 0.073  & 0.074  & 0.080  & 0.075 \\
        & FreeControl &  0.119  & 0.067  & 0.144  & 0.143  & 0.106  & 0.097  & 0.113 \\ 
        & DriveGEN &  \textbf{0.641}  & \textbf{0.616}  & \textbf{0.632}  & \textbf{0.633}  & \textbf{0.614}  & \textbf{0.627}  & \textbf{0.627} \\         
         \bottomrule
         \end{tabular}
         }
    \end{center}
    % \vspace{-0.25in}
\end{table}

\noindent
\textbf{More Discussions on Image Quality.}
We first report PSNR and SSIM for object regions between generated and original images. Table~\ref{tab:PSNR_SSIM} reveals \ournet~stably preserves object geometries within all scenarios without compromising the quality.
Note that preserving object information is crucial for reusing annotations, as misaligned objects may introduce bias as shown in Figure \textbf{\textcolor{Red}{2}} in the manuscript.

\begin{table}[!h]
\setlength\tabcolsep{3pt}
    \begin{center}
    % \vspace{-0.1in}
    \caption{\label{tab:steps} 
    Comparisons of the detector performance enhanced by \ournet~with various diffusion steps.}
    \scalebox{0.6}{
         \begin{tabular}{c|c|c|cccccccc}
         \toprule
         KITTI-C & MonoFlex & FreeControl(200 steps) & \ournet-50 & \ournet-100 & \ournet-200 \\
         \midrule
          Time (hour) & 36.85  & 43.97  & \textbf{8.69}  & 32.40  & 52.97  \\
          Car  (mAP) & 26.45  & 22.44  & 32.48  & 33.93  & \textbf{34.07}  \\
         \bottomrule
         \end{tabular}
         }
    \end{center}
    % \vspace{-0.28in}
\end{table}

%-------------------------------------------------------------------------------------------

%-------------------------------------------------------------------------------------------

%-------------------------------------------------------------------------------------------
\begin{figure*}[t] 
  \centering
  \includegraphics[width=\linewidth]{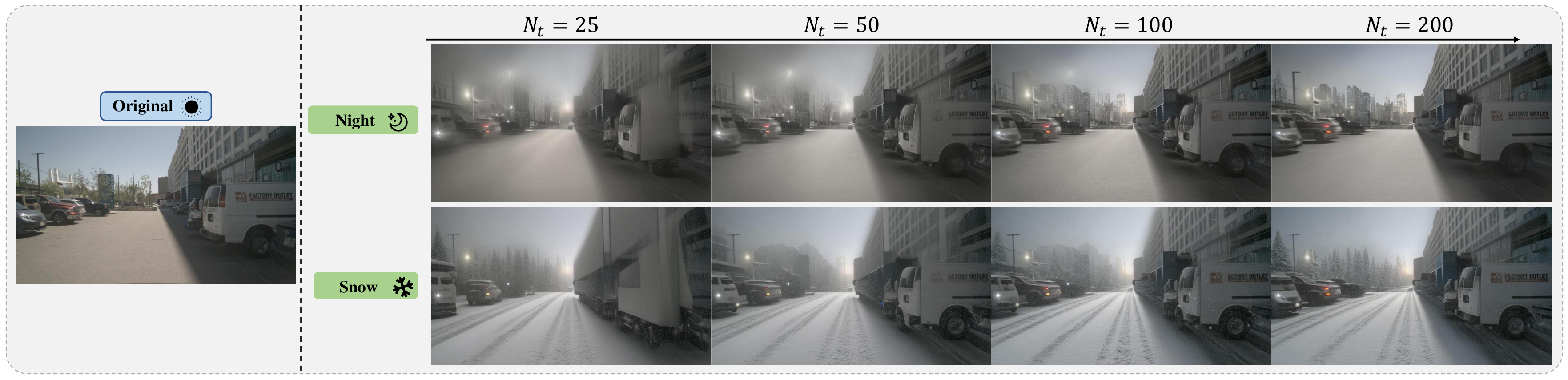} 
  \caption{More qualitative results of DriveGEN with varying numbers of diffusion steps.}
  \label{fig:vis_qua}
\end{figure*}
% \clearpage
%-------------------------------------------------------------------------------------------

Besides, to explore the computation cost and improve the efficiency, we provide further analysis of \ournet~with different diffusion step settings. As shown in Table~\ref{tab:steps}, we conduct experiments with 50 and 100 generation steps (8$\times$3090 GPUs). With fewer steps, \ournet~drastically reduces time consumption while still achieving sufficient gains. 
To further validate \ournet, we also present additional qualitative results obtained with varying numbers of diffusion steps as shown in Figure~\ref{fig:vis_qua}.
It is observed that \ournet~consistently preserves objects even when only 50 steps ($N_t=50$) are used. Moreover, increasing the number of steps leads to progressively enhanced image quality.
These results reveal the inherent trade-off between image quality and efficiency, thereby enabling users to tailor the approach according to their application demands.

% As \(\tau\) increases, the corresponding corruption progressively exerts a more severe impact on the background; however, the objects remain well-preserved.
% the robustness of detectors progressively improves with the increasing of additional augmented scenarios

% \clearpage

\clearpage
\begin{table*}[h]
\setlength\tabcolsep{7pt}
\renewcommand\arraystretch{1.0}
    \begin{center}
    \caption{\label{tab:kitti-c-moderate} Comparison on the {KITTI-C} dataset, severity \textbf{level 1} regarding {\textbf{Moderate} $AP_{3D|R_{40}}$}. The \textbf{bold} number indicates the best result. 
    }  
    % \vspace{-0.08in}
    \scalebox{0.55}{
         \begin{tabular}{lc|ccc|ccc|cccc|ccc|c}
         \toprule
         \multicolumn{15}{c}{\textbf{\textbf{Car}, IoU @ 0.7, 0.5, 0.5}} \\
         \midrule
         \multirow{2}{*}{Method}  &
         \multirow{2}{*}{\shortstack{Training-free\\diffusion}} &
         \multicolumn{3}{c|}{Noise} & 
         \multicolumn{3}{c|}{Blur} & 
         \multicolumn{4}{c|}{Weather} & 
         \multicolumn{3}{c|}{Digital} & 
         \multirow{2}{*}{Avg.} \\
        \cmidrule(lr){3-5} \cmidrule(lr){6-8} \cmidrule(lr){9-12} \cmidrule(lr){13-15} 
        & & Gauss. & Shot & Impul. & Defoc. & Glass & Motion & Snow & Frost & Fog & Brit. & Contr. & Pixel & Sat. \\
         \midrule

        Monoflex~\cite{zhang2021objects} &  & 12.69 & 19.65 & 13.88 & 17.81 & 25.86 & 27.44 & 31.53 & 28.77 & 18.90 & 42.36 & 18.94 & 26.48 & 35.51 & 24.60 \\
        \midrule
        
        ~$\bullet~$ Color Jitter (Traditional aug.) & & 9.75 & 15.31 & 11.84 & 20.55 & 23.02 & 27.19 & 31.55 & 28.69 & 18.62 & 39.22 & 19.32 & 15.85 & 32.95 & 22.61  \\
        ~$\bullet~$ Brightness (Traditional aug.) & & 10.76 & 17.60 & 11.99 & 10.06 & 17.14 & 19.25 & 25.22 & 20.77 & 12.55 & 36.73 & 12.38 & 19.35 & 27.79 & 18.58  \\
        
        \midrule
        ~$\bullet~$ ControlNet (Only Snow aug.) & \xmark &  0.33 & 1.21 & 1.44 & 3.99 & 10.07 & 15.37 & 21.99 & 19.16 & 9.20 & 32.35 & 8.86 & 1.30 & 16.00 & 10.87 \\  
        ~$\bullet~$ ControlNet (3 scenarios aug.) & \xmark &  1.01 & 1.10 & 0.48 & 0.39 & 0.41 & 0.84 & 4.04 & 3.20 & 1.24 & 9.03 & 1.11 & 0.46 & 3.60 & 2.07  \\  
        ~$\bullet~$ ControlNet (6 scenarios aug.) & \xmark & 0.00 & 0.00 & 0.00 & 0.00 & 0.00 & 1.88 & 0.00 & 0.00 & 0.00 & 0.00 & 0.00 & 0.00 & 0.00 & 0.14  \\  
         \midrule
        ~$\bullet~$ Freecontrol (Only Snow aug.) & \cmark &  18.31 & 25.11 & 19.97 & 11.37 & 19.31 & 18.88 & 25.29 & 16.56 & 9.30 & 31.85 & 9.13 & 25.37 & 31.73 & 20.17  \\  
        ~$\bullet~$ Freecontrol (3 scenarios aug.) & \cmark & 13.39 & 18.37 & 11.69 & 14.06 & 17.39 & 14.96 & 19.65 & 15.63 & 14.17 & 21.49 & 15.40 & 22.21 & 27.72 & 17.39 \\  
        ~$\bullet~$ Freecontrol (6 scenarios aug.) & \cmark & 11.48 & 16.64 & 14.79 & 11.25 & 14.73 & 11.51 & 16.06 & 14.93 & 13.18 & 21.69 & 13.83 & 20.03 & 21.74 & 15.53 \\  
         \midrule
        ~$\bullet~$ DriveGEN (Only Snow aug.) & \cmark & 15.82 & 24.72 & 23.33 & 28.17 & 32.42 & 34.85 & 38.00 & 35.39 & 25.90 & \textbf{45.99} & 27.51 & 35.55 & \textbf{40.41} & 31.39  \\  
        ~$\bullet~$ DriveGEN (3 scenarios aug.) & \cmark &  \textbf{23.71} & \textbf{33.55} & \textbf{26.80} & 30.26 & 35.20 & 33.15 & 35.60 & 34.68 & 31.59 & 41.10 & 31.70 & 35.50 & 39.92 & 33.29  \\  
        ~$\bullet~$ DriveGEN (6 scenarios aug.) & \cmark & 23.24 & 31.24 & 26.40 & \textbf{33.48} & \textbf{36.20} & \textbf{37.00} & \textbf{37.21} & \textbf{37.10} & \textbf{35.16} & 41.00 & \textbf{35.93} & \textbf{38.92} & 40.30 & \textbf{34.86}  \\ 
        
        \midrule
        \midrule

        MonoGround~\cite{qin2022monoground} &  & 12.30 & 20.27 & 17.08 & 18.41 & 27.22 & 29.04 & 32.16 & 25.56 & 13.34 & 43.41 & 14.14 & 31.21 & 32.55 & 24.36  \\
        \midrule
        
        ~$\bullet~$ Color Jitter (Traditional aug.) & &  11.96 & 22.17 & 17.69 & 20.67 & 27.17 & 28.47 & 32.93 & 28.37 & 19.10 & 41.85 & 19.58 & 27.59 & 33.42 & 25.46  \\
        ~$\bullet~$ Brightness (Traditional aug.) & & 13.08 & 21.53 & 18.18 & 21.98 & 28.58 & 26.12 & 32.29 & 30.67 & 17.75 & 39.47 & 16.92 & 24.05 & 34.25 & 24.99  \\
        
        \midrule
        ~$\bullet~$ ControlNet (Only Snow aug.) & \xmark &  1.81 & 2.72 & 4.08 & 5.02 & 12.18 & 13.18 & 16.51 & 11.40 & 2.68 & 33.23 & 2.69 & 6.52 & 12.53 & 9.58 \\  
        ~$\bullet~$ ControlNet (3 scenarios aug.) & \xmark &  0.00 & 0.00 & 0.27 & 1.49 & 1.76 & 1.86 & 5.16 & 5.29 & 0.59 & 17.31 & 1.29 & 7.63 & 6.00 & 3.74  \\  
        ~$\bullet~$ ControlNet (6 scenarios aug.) & \xmark & 0.00 & 0.00 & 0.00 & 1.83 & 1.05 & 0.40 & 1.44 & 0.53 & 0.45 & 3.47 & 0.38 & 2.17 & 1.52 & 1.02  \\  
         \midrule
        ~$\bullet~$ Freecontrol (Only Snow aug.) & \cmark &  11.09 & 19.54 & 14.50 & 15.84 & 18.97 & 19.18 & 29.09 & 19.34 & 13.10 & 32.39 & 13.07 & 24.27 & 34.87 & 20.40  \\  
        ~$\bullet~$ Freecontrol (3 scenarios aug.) & \cmark & 14.22 & 18.36 & 15.44 & 11.34 & 14.79 & 12.63 & 15.89 & 14.80 & 11.07 & 22.18 & 12.62 & 19.93 & 21.82 & 15.78 \\  
        ~$\bullet~$ Freecontrol (6 scenarios aug.) & \cmark & 13.30 & 19.87 & 13.29 & 19.34 & 18.40 & 16.58 & 15.59 & 13.14 & 13.22 & 21.52 & 15.49 & 20.72 & 23.59 & 17.24 \\  
         \midrule
        ~$\bullet~$ DriveGEN (Only Snow aug.) & \cmark &  15.53 & 24.35 & 21.67 & 29.67 & 33.88 & 35.40 & \textbf{37.81} & 32.55 & 23.75 & \textbf{43.15} & 25.75 & 34.80 & 41.17 & 30.73 \\  
        ~$\bullet~$ DriveGEN (3 scenarios aug.) & \cmark &  18.12 & 29.12 & 25.30 & 33.18 & 36.06 & 35.28 & 36.08 & 33.67 & 26.57 & 41.40 & 27.03 & 39.13 & \textbf{41.57} & 32.50  \\  
        ~$\bullet~$ DriveGEN (6 scenarios aug.) & \cmark & \textbf{21.78} & \textbf{29.21} & \textbf{27.16} & \textbf{34.57} & \textbf{36.52} & \textbf{35.91} & 34.70 & \textbf{34.91} & \textbf{29.74} & 39.94 & \textbf{31.94} & \textbf{40.06} & 40.69 & \textbf{33.63} \\ 

        \midrule
        \midrule
        \multicolumn{15}{c}{\textbf{\textbf{Pedestrian}, IoU @ 0.7, 0.5, 0.5}} \\
        \midrule
        \midrule
        
        Monoflex~\cite{zhang2021objects} &  &  0.98 & 3.79 & 0.81 & 7.91 & 17.25 & 14.60 & 13.23 & 9.09 & 4.99 & 19.51 & 5.34 & 1.63 & 8.64 & 8.29 \\
        \midrule
        
        ~$\bullet~$ Color Jitter (Traditional aug.) & & 0.88 & 3.59 & 1.61 & 9.86 & 14.48 & 12.12 & 14.86 & 11.94 & 7.51 & 18.39 & 9.29 & 0.97 & 11.11 & 8.97  \\
        ~$\bullet~$ Brightness (Traditional aug.) & & 0.79 & 1.88 & 1.08 & 4.80 & 13.74 & 12.14 & 7.54 & 5.85 & 1.70 & 16.75 & 2.57 & 0.62 & 3.69 & 5.62  \\
        
        \midrule
        ~$\bullet~$ ControlNet (Only Snow aug.) & \xmark & 0.00 & 0.00 & 0.00 & 1.74 & 7.92 & 5.23 & 3.62 & 3.52 & 1.41 & 10.79 & 1.54 & 0.00 & 0.66 & 2.80   \\  
        ~$\bullet~$ ControlNet (3 scenarios aug.) & \xmark &  0.00 & 0.00 & 0.00 & 1.19 & 2.78 & 1.15 & 1.67 & 1.32 & 0.83 & 4.11 & 2.05 & 0.00 & 0.00 & 1.16  \\  
        ~$\bullet~$ ControlNet (6 scenarios aug.) & \xmark & 0.00 & 0.00 & 0.00 & 0.00 & 0.36 & 0.00 & 0.00 & 0.00 & 0.00 & 2.50 & 0.00 & 0.00 & 0.00 & 0.22 \\  
         \midrule
        ~$\bullet~$ Freecontrol (Only Snow aug.) & \cmark & 3.87 & 5.18 & 3.52 & 3.20 & 5.07 & 5.66 & 4.75 & 3.44 & 3.94 & 7.42 & 3.77 & 8.07 & 7.44 & 5.03  \\  
        ~$\bullet~$ Freecontrol (3 scenarios aug.) & \cmark & 2.69 & 3.74 & 3.06 & 3.07 & 5.01 & 4.22 & 7.44 & 4.62 & 7.18 & 10.45 & 6.07 & 6.82 & 7.28 & 5.51 \\  
        ~$\bullet~$ Freecontrol (6 scenarios aug.) & \cmark & 5.17 & 8.71 & 8.56 & 4.18 & 6.82 & 5.08 & 6.98 & 7.18 & 6.35 & 10.19 & 9.18 & 9.45 & 9.32 & 7.47  \\  
         \midrule
        ~$\bullet~$ DriveGEN (Only Snow aug.)   & \cmark &  1.10 & 3.17 & 4.32 & 16.14 & \textbf{19.36} & \textbf{20.31} & \textbf{19.45} & 14.44 & 8.77 & \textbf{24.07} & 8.86 & 9.59 & 17.64 & 12.86 \\  
        ~$\bullet~$ DriveGEN (3 scenarios aug.) & \cmark &  5.60 & \textbf{9.66} & \textbf{9.01} & 14.73 & 18.29 & 16.45 & 15.30 & \textbf{15.81} & 15.30 & 21.73 & 16.81 & 13.84 & \textbf{17.73} & 14.63  \\  
        ~$\bullet~$ DriveGEN (6 scenarios aug.) & \cmark &  \textbf{6.44} & 9.32 & 7.10 & \textbf{17.46} & 19.25 & 19.89 & 16.89 & 15.51 & \textbf{15.50} & 23.48 & \textbf{16.64} & \textbf{14.37} & 17.12 & \textbf{15.31} \\ 
        
        \midrule
        \midrule
        
        MonoGround~\cite{qin2022monoground} &  & 2.81 & 3.40 & 5.76 & 17.46 & 18.40 & 17.40 & 12.64 & 9.06 & 4.01 & 23.33 & 5.51 & 3.08 & 7.06 & 9.99 \\
        \midrule
        
        ~$\bullet~$ Color Jitter (Traditional aug.) & & 2.44 & 3.35 & 3.67 & 14.85 & 18.04 & 15.69 & 14.63 & 12.06 & 9.01 & 24.26 & 9.34 & 2.06 & 7.76 & 10.55  \\
        ~$\bullet~$ Brightness (Traditional aug.) & & 2.85 & 4.23 & 7.50 & 13.84 & 13.78 & 13.96 & 11.52 & 12.34 & 5.60 & 20.39 & 5.27 & 2.25 & 10.40 & 9.53  \\
        
        \midrule
        ~$\bullet~$ ControlNet (Only Snow aug.) & \xmark & 1.88 & 1.03 & 0.80 & 7.44 & 9.68 & 8.30 & 0.91 & 2.93 & 1.25 & 12.78 & 1.42 & 0.26 & 0.96 & 3.82   \\  
        ~$\bullet~$ ControlNet (3 scenarios aug.) & \xmark &  0.00 & 0.00 & 0.00 & 3.32 & 3.85 & 2.32 & 1.17 & 1.05 & 1.50 & 4.69 & 1.25 & 0.26 & 0.77 & 1.55  \\  
        ~$\bullet~$ ControlNet (6 scenarios aug.) & \xmark & 0.00 & 0.00 & 0.00 & 0.00 & 0.00 & 2.50 & 0.00 & 0.00 & 0.00 & 0.00 & 0.00 & 0.00 & 0.00 & 0.19 \\  
         \midrule
        ~$\bullet~$ Freecontrol (Only Snow aug.) & \cmark & \textbf{9.63} & 12.34 & 10.78 & 14.08 & 12.11 & 12.79 & 15.08 & 12.35 & 7.72 & 15.06 & 11.47 & 15.99 & 14.16 & 12.58   \\  
        ~$\bullet~$ Freecontrol (3 scenarios aug.) & \cmark & 0.82 & 1.80 & 1.84 & 2.30 & 3.14 & 3.77 & 3.75 & 0.40 & 2.84 & 3.73 & 3.43 & 3.78 & 5.49 & 2.85 \\  
        ~$\bullet~$ Freecontrol (6 scenarios aug.) & \cmark & 6.48 & 7.49 & 7.75 & 5.51 & 5.44 & 4.85 & 2.72 & 1.36 & 4.16 & 8.23 & 4.88 & 9.02 & 6.30 & 5.71 \\  
         \midrule
        ~$\bullet~$ DriveGEN (Only Snow aug.) & \cmark & 6.11 & 6.69 & 8.96 & 14.75 & 16.80 & 18.38 & 16.23 & 12.99 & 9.64 & 22.73 & 11.25 & 11.78 & 14.07 & 13.11  \\  
        ~$\bullet~$ DriveGEN (3 scenarios aug.) & \cmark &  6.53 & 9.20 & 10.90 & 15.78 & \textbf{19.10} & \textbf{20.51} & \textbf{18.01} & 15.79 & 12.98 & \textbf{24.46} & 13.92 & 19.26 & \textbf{19.29} & 15.83  \\  
        ~$\bullet~$ DriveGEN (6 scenarios aug.) & \cmark & 9.49 & \textbf{13.28} & \textbf{13.02} & \textbf{16.33} & 17.10 & 19.67 & 16.70 & \textbf{18.02} & \textbf{15.35} & 22.42 & \textbf{14.99} & \textbf{19.72} & 17.80 & \textbf{16.45}  \\ 

        \midrule
        \midrule
        \multicolumn{15}{c}{\textbf{\textbf{Cyclist}, IoU @ 0.7, 0.5, 0.5}} \\
        \midrule
        \midrule        
        Monoflex~\cite{zhang2021objects} &  & 0.48 & 1.59 & 0.53 & 2.05 & 6.41 & 7.45 & \textbf{9.93} & \textbf{8.61} & 4.70 & \textbf{14.08} & 3.54 & 2.80 & 7.82 & 5.38 \\
        \midrule
        
        ~$\bullet~$ Color Jitter (Traditional aug.) & & 0.52 & 2.45 & \textbf{1.51} & 1.29 & 2.79 & 6.36 & 8.72 & 8.08 & 3.55 & 10.00 & 3.62 & 0.84 & 7.92 & 4.43  \\
        ~$\bullet~$ Brightness (Traditional aug.) & & 0.16 & 0.83 & 0.19 & 1.26 & 2.63 & 4.81 & 7.67 & 6.91 & 3.64 & 10.67 & 4.30 & 1.68 & 6.22 & 3.92 \\
        
        \midrule
        ~$\bullet~$ ControlNet (Only Snow aug.) & \xmark &  0.00 & 0.26 & 0.00 & 0.00 & 3.13 & 3.20 & 5.95 & 4.85 & 5.67 & 12.72 & 4.88 & 1.67 & 1.68 & 3.39 \\  
        ~$\bullet~$ ControlNet (3 scenarios aug.) & \xmark &  0.00 & 0.00 & 0.00 & 0.00 & 0.00 & 0.00 & 0.00 & 2.50 & 1.50 & 1.58 & 1.35 & 0.00 & 0.00 & 0.53  \\  
        ~$\bullet~$ ControlNet (6 scenarios aug.) & \xmark & 0.00 & 0.00 & 0.00 & 0.00 & 0.00 & 0.00 & 0.00 & 0.00 & 0.00 & 0.00 & 0.00 & 0.00 & 0.00 & 0.00 \\  
         \midrule
        ~$\bullet~$ Freecontrol (Only Snow aug.) & \cmark & 1.13 & 3.22 & 1.34 & 0.00 & 0.39 & 0.67 & 2.89 & 1.06 & 0.29 & 4.07 & 0.33 & 0.65 & 6.16 & 1.71  \\  
        ~$\bullet~$ Freecontrol (3 scenarios aug.) & \cmark & 0.00 & 0.00 & 0.00 & 0.00 & 2.50 & 0.00 & 0.91 & 1.04 & 1.43 & 3.03 & 1.33 & 0.00 & 2.05 & 0.95  \\  
        ~$\bullet~$ Freecontrol (6 scenarios aug.) & \cmark & 0.20 & 0.18 & 0.22 & 0.00 & 0.31 & 0.00 & 0.52 & 0.52 & 0.76 & 1.61 & 1.82 & 0.60 & 0.61 & 0.57 \\  
         \midrule
        ~$\bullet~$ DriveGEN (Only Snow aug.) & \cmark & 0.43 & 0.99 & 0.50 & 1.02 & 4.09 & 4.70 & 8.48 & 5.43 & 2.69 & 12.12 & 3.35 & 1.42 & 8.52 & 4.13 \\  
        ~$\bullet~$ DriveGEN (3 scenarios aug.) & \cmark &  \textbf{0.87} & \textbf{2.70} & 1.07 & 1.66 & 4.01 & 5.98 & 6.22 & 2.87 & 3.72 & 10.14 & 3.53 & 3.59 & 8.40 & 4.21  \\  
        ~$\bullet~$ DriveGEN (6 scenarios aug.) & \cmark & 0.50 & 1.62 & 0.64 & \textbf{3.06} & \textbf{6.00} & \textbf{7.73} & 7.46 & 8.11 & \textbf{8.14} & 11.55 & \textbf{9.11} & \textbf{4.16} & \textbf{8.82} & \textbf{5.92}  \\ 
        
        \midrule
        \midrule
        
        MonoGround~\cite{qin2022monoground} &  & 0.13 & 1.39 & 1.01 & 0.63 & 2.51 & 1.82 & 3.90 & 2.81 & 0.66 & 8.91 & 0.90 & 1.84 & 4.33 & 2.37 \\
        \midrule
        
        ~$\bullet~$ Color Jitter (Traditional aug.) & &  0.30 & \textbf{2.00} & \textbf{1.59} & 0.25 & 1.60 & 1.73 & 4.00 & 3.60 & 0.74 & \textbf{10.33} & 1.30 & 2.19 & 3.96 & 2.58 \\
        ~$\bullet~$ Brightness (Traditional aug.) & & 0.04 & 0.47 & 0.17 & 0.36 & 1.05 & 0.84 & 3.67 & 1.69 & 1.12 & 5.30 & 0.55 & 0.67 & 2.15 & 1.39  \\
        
        \midrule
        ~$\bullet~$ ControlNet (Only Snow aug.) & \xmark &  0.00 & 0.00 & 0.67 & 0.00 & 0.56 & 1.25 & 0.39 & 0.44 & 0.12 & 5.12 & 0.28 & 0.37 & 0.51 & 0.75  \\  
        ~$\bullet~$ ControlNet (3 scenarios aug.) & \xmark &  0.00 & 0.00 & 0.00 & 0.00 & 0.30 & 0.00 & 0.83 & 1.50 & 0.00 & 1.22 & 0.00 & 0.31 & 0.00 & 0.32  \\  
        ~$\bullet~$ ControlNet (6 scenarios aug.) & \xmark & 0.00 & 0.00 & 0.00 & 0.00 & 0.00 & 0.00 & 0.00 & 0.00 & 0.00 & 0.00 & 0.00 & 0.00 & 0.00 & 0.00 \\  
         \midrule
        ~$\bullet~$ Freecontrol (Only Snow aug.) & \cmark &  0.33 & 0.42 & 0.29 & 1.07 & 0.16 & 0.26 & 1.42 & 0.63 & 0.20 & 3.59 & 0.17 & 1.34 & 3.65 & 1.04  \\  
        ~$\bullet~$ Freecontrol (3 scenarios aug.) & \cmark & 0.14 & 0.37 & 0.32 & 0.00 & 0.37 & 0.50 & 0.37 & 0.33 & 0.58 & 1.09 & 0.42 & 0.27 & 0.29 & 0.39 \\  
        ~$\bullet~$ Freecontrol (6 scenarios aug.) & \cmark & 0.00 & 0.32 & 0.65 & 0.00 & 0.42 & 0.54 & 0.00 & 0.00 & 1.00 & 1.70 & 0.83 & 1.25 & 0.74 & 0.57 \\  
         \midrule
        ~$\bullet~$ DriveGEN (Only Snow aug.) & \cmark & 0.10 & 0.54 & 0.29 & 0.31 & 1.45 & 2.89 & 2.92 & 1.48 & 0.74 & 6.44 & 1.04 & 1.39 & 3.66 & 1.79  \\  
        ~$\bullet~$ DriveGEN (3 scenarios aug.) & \cmark & \textbf{0.43} & 1.41 & 1.12 & 0.80 & 2.37 & 2.10 & 4.13 & 4.12 & 0.78 & 7.71 & 0.85 & \textbf{3.22} & \textbf{4.72} & 2.60   \\  
        ~$\bullet~$ DriveGEN (6 scenarios aug.) & \cmark & 1.17 & 1.52 & 1.25 & \textbf{2.88} & \textbf{4.62} & \textbf{4.25} & \textbf{4.54} & \textbf{4.37} & \textbf{1.79} & 7.43 & \textbf{2.65} & 2.63 & 4.17 & \textbf{3.33}  \\ 

         \bottomrule
         \end{tabular}
         }
    \end{center}
\end{table*}

\clearpage


\end{document}